\documentclass[10pt,twocolumn,letterpaper]{article}

\usepackage[pagenumbers]{cvpr} % To force page numbers, e.g. for an arXiv version

\usepackage{graphicx}
\usepackage{amsmath}
\usepackage{amssymb}
\usepackage{booktabs}
\usepackage{wasysym}

\usepackage[pagebackref,breaklinks,colorlinks]{hyperref}
\usepackage{url}
\usepackage{xspace}
\usepackage{enumitem}
\usepackage[table]{xcolor}
\usepackage{booktabs}
\usepackage{multicol}
\usepackage{multirow}
\usepackage{graphicx}
\usepackage{array}

\newcolumntype{?}{!{\vrule width 1pt}}
\newcommand{\mypara}{\vspace{1mm}\noindent\textbf}
\newcommand{\hab}{\textsc{Habitat}\xspace}
\newcommand{\robo}{\textsc{RoboTHOR}\xspace}
\newcommand{\dataset}{\textsc{Pasture}\xspace}

\newcommand{\customfootnotetext}[2]{{% Group to localize change to footnote
  \renewcommand{\thefootnote}{#1}% Update footnote counter representation
  \footnotetext[0]{#2}}}

\definecolor{tab:pink}{RGB}{213,126,190}
\definecolor{tab:orange}{RGB}{239,133,54}
\definecolor{tab:green}{RGB}{81,147,62}
\definecolor{tab:purple}{RGB}{141,106,184}
\definecolor{tab:cyan}{RGB}{87,187,204}

\definecolor{LightCyan}{rgb}{0.88,1,1}
\definecolor{yel}{rgb}{1,1,0.88}

\newcolumntype{a}{>{\columncolor{LightCyan}}c}
\newcolumntype{y}{>{\columncolor{yel}}c}

\usepackage[capitalize]{cleveref}
\crefname{section}{Sec.}{Secs.}
\Crefname{section}{Section}{Sections}
\Crefname{table}{Table}{Tables}
\crefname{table}{Tab.}{Tabs.}

\def\cvprPaperID{4393} % *** Enter the CVPR Paper ID here
\def\confName{CVPR}
\def\confYear{2023}

\begin{document}

% \title{CLIP on Wheels:\\Open-vocabulary models are (almost) zero-shot object navigators}
\title{CoWs on \dataset: Baselines and Benchmarks\\ for Language-Driven Zero-Shot Object Navigation}

\date{}
\author{
        Samir Yitzhak Gadre$^\diamond$\quad
        Mitchell Wortsman$^{\dag}$\quad
        Gabriel Ilharco$^{\dag}$\quad
        Ludwig Schmidt$^{\dag}$\quad
        Shuran Song$^\diamond$
}

\maketitle

\customfootnotetext{$\diamond$}{
Columbia University, $^{\dag}$University of Washington.
Correspondence to \texttt{sy@cs.columbia.edu}.}

%%%%%%%%% ABSTRACT
\begin{abstract}

For robots to be generally useful, they must be able to find arbitrary objects described by people (i.e., be language-driven) even without expensive navigation training on in-domain data (i.e., perform zero-shot inference). We explore these capabilities in a unified setting: language-driven zero-shot object navigation (L-ZSON). Inspired by the recent success of open-vocabulary models for image classification, we investigate a straightforward framework, CLIP on Wheels (CoW), to adapt open-vocabulary models to this task without fine-tuning. To better evaluate L-ZSON, we introduce the \dataset benchmark, which considers finding uncommon objects, objects described by spatial and appearance attributes, and hidden objects described relative to visible objects. We conduct an in-depth empirical study by directly deploying 21 CoW baselines across \hab, \robo, and \dataset. In total, we evaluate over 90k navigation episodes and find that (1) CoW baselines often struggle to leverage language descriptions, but are proficient at finding uncommon objects. (2) A simple CoW, with CLIP-based object localization and classical exploration---and no additional training---matches the navigation efficiency of a state-of-the-art ZSON method trained for 500M steps on \hab MP3D data. This same CoW provides a 15.6 percentage point improvement in success over a state-of-the-art \robo ZSON model.\footnote{For code, data, and videos, see \href{https://cow.cs.columbia.edu/}{\texttt{cow.cs.columbia.edu/}}}

\end{abstract}

%%%%%%%%% BODY TEXT
\section{Introduction}
\label{sec:intro}
To be more widely applicable, robots should be language-driven: able to deduce goals based on arbitrary text input instead of being constrained to a fixed set of object categories.
While existing image classification, semantic segmentation, and object navigation benchmarks like ImageNet-1k~\cite{russakovsky_2015}, ImageNet-21k~\cite{Deng2009ImageNetAL}, MS-COCO~\cite{lin_2015}, LVIS~\cite{Gupta2019LVISAD}, \hab~\cite{habitat19iccv}, and \robo~\cite{Deitke2020RoboTHOR} include a vast array of everyday items, they do not capture all objects that matter to people.
For instance, a lost ``toy airplane" may become relevant in a kindergarten classroom, but this object is not annotated in any of the aforementioned datasets.

In this paper, we study \emph{Language-driven zero-shot object navigation (L-ZSON)}---a more challenging but also more applicable version of object navigation~\cite{zhu2017target,wortsman2019learning,Batra2020ObjectNavRO,habitat19iccv,Deitke2020RoboTHOR} and ZSON~\cite{Khandelwal2021Simple,majumdar2022zson} tasks.
In L-ZSON, an agent must find an object based on a natural language description, which may contain different levels of granularity (e.g., ``toy airplane", ``toy airplane under the bed", or ``wooden toy airplane").
L-ZSON encompasses ZSON, which specifies only the target category~\cite{Khandelwal2021Simple,majumdar2022zson}.
Since L-ZSON is ``zero-shot", we consider agents that do not have access to navigation training on the target objects or domains.
This reflects realistic application scenarios, where the environment and object set may not be known a priori.

\begin{figure}[tp]
    \centering
    \includegraphics[width=\linewidth]{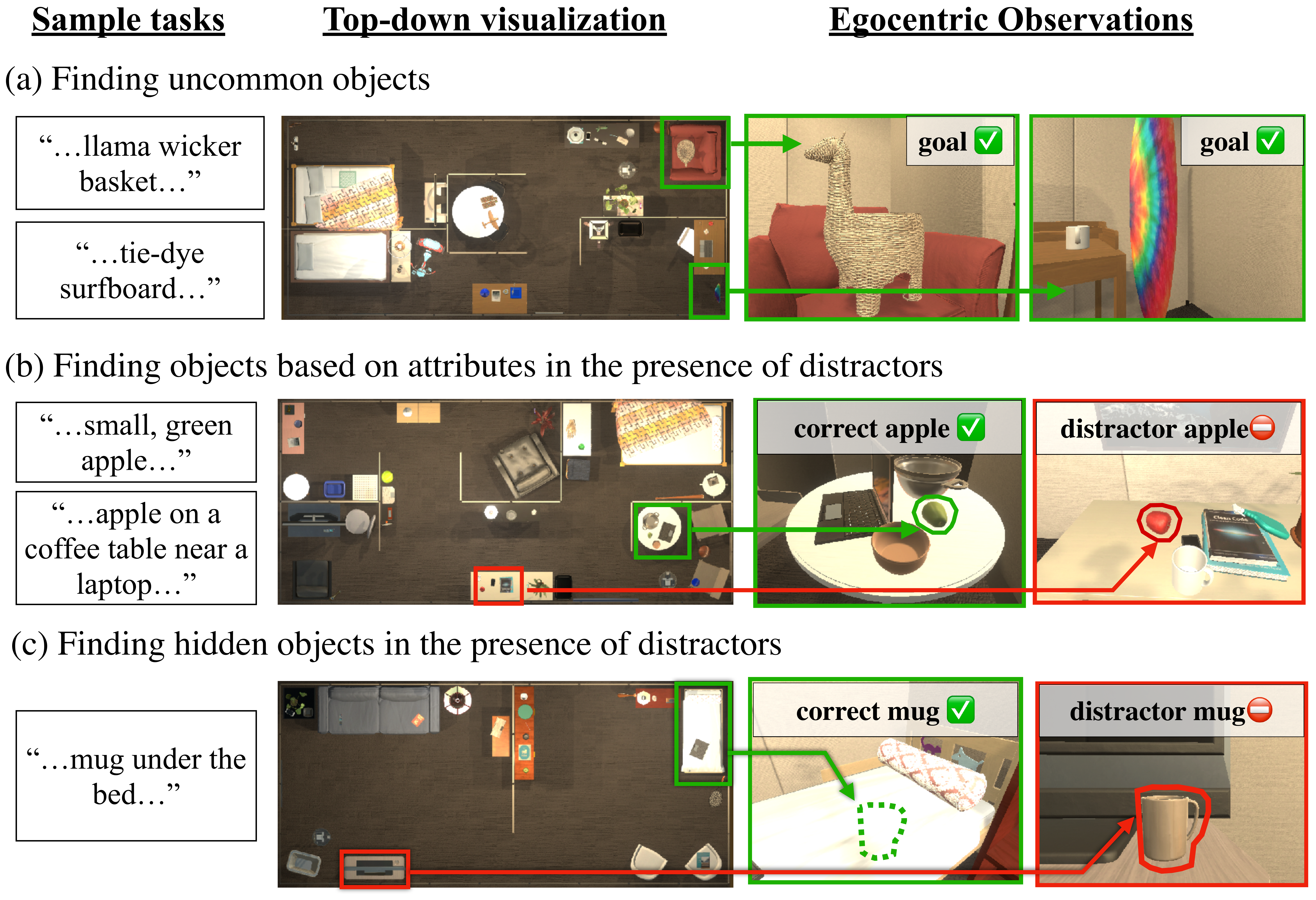}
    \vspace*{-5mm}
    \caption{\textbf{The \dataset benchmark for L-ZSON.}
    Text specifies navigation goal objects.
    Agents do not train on these tasks, making the evaluation protocol zero-shot.
    (a) \emph{Uncommon object goals} like ``llama wicker basket", not found in existing navigation benchmarks.
    (b) \emph{Appearance, spatial descriptions}, which are necessary to find the correct object.
    (c) \emph{Hidden object descriptions}, which localize objects that are not visible.
    }
    \vspace{-5mm}
    \label{fig:teaser}
\end{figure}

Performing L-ZSON in \emph{any} environment with \emph{unstructured} language input is challenging; however, recent advances in open-vocabulary models for image classification~\cite{Radford2021LearningTV,jia2021scaling, pham2021scaling}, object detection \cite{bansal2018zero,Rahman2018ZeroShotOD,Demirel2018ZeroShotOD,Li2019ZeroShotOD,Zhu2020ZeroSD,Mao2020ZeroShotOD,Gu2021ZeroShotDV,Kamath2021MDETRM,Minderer2022SimpleOO}, and semantic segmentation \cite{NEURIPS2019_0266e33d,Kato_2019_ICCV,NEURIPS2020_f73b76ce,Baek_2021_ICCV,Cheng_2021_ICCV,Kamath2021MDETRM,Zhou2021DenseCLIPEF} present a promising foundation.
These models provide an interface where one specifies---in text---the arbitrary objects they wish to classify, detect, or segment.
For example, CLIP~\cite{Radford2021LearningTV} open-vocabulary classifiers compute similarity scores between an input image and a set of user-specified captions (e.g., ``a photo of a toy airplane.", ...), selecting the caption with the highest score to determine the image classification label.
Given the flexibility of these models, we would like to understand their capability to execute embodied tasks even without additional training.

We present baselines and benchmarks for L-ZSON. More specifically: 
\begin{itemize}[leftmargin=3mm]
    \vspace{-2mm}
    \item \emph{A collection of baseline algorithms, CLIP on Wheels (CoW), which adapt open-vocabulary models to the task of L-ZSON}. CoW takes inspiration from the semantic mapping line of work~\cite{kuipers1991robot,Nicholson2019QuadricSLAMDQ,Chaplot2020LearningTE}, and decomposes the navigation task into exploration when the language target is not confidently localized, and target-driven planning otherwise. CoW retains the textual user interface of the original open-vocabulary model and does not require any navigation training. We evaluate 21 CoWs, ablating over many open-vocabulary models, exploration policies, backbones, prompting strategies, and post-processing strategies.
    \vspace{-2mm}
    \item \emph{A new benchmark, \dataset, to evaluate CoW and future methods on L-ZSON}. We design \dataset, visualized in Fig.~\ref{fig:teaser}, to study capabilities that traditional object navigation agents, which are trained on a fixed set of categories, do not possess.
    We consider the ability to find (1) uncommon objects (e.g., ``tie-dye surfboard"), (2) objects by their spatial and appearance attributes in the presence of distractor objects (e.g., ``green apple" vs. ``red apple"), and (3) objects that cannot be visually observed (e.g., ``mug under the bed").
    \vspace{-1mm}
\end{itemize}

Together the CoW baselines and \dataset benchmark allow us to conduct extensive studies on the capabilities of open-vocabulary models in the context of L-ZSON embodied tasks.
Our experiments demonstrate CoW's potential on uncommon objects and limitations in taking full advantage of language descriptions---thereby providing empirical motivation for future studies.
To contextualize CoW relative to prior zero-shot methods, we additionally evaluate on the \hab MP3D dataset. We find that our best CoW achieves navigation efficiency (\textsc{SPL}) that matches a state-of-the-art ZSON method~\cite{majumdar2022zson} that trains on MP3D training data for 500M steps.
On a \robo object subset, considered in prior work, the same CoW beats the leading method~\cite{Khandelwal2021Simple} by 15.6 percentage points in task success.
\section{Related Work}
\label{sec:related}
\mypara{Mapping and exploration.}
Exploring effectively with a mobile robot is a long-standing problem in vision and robotics.  
Classical methods often decompose the task into map reconstruction~\cite{henry2014rgb,moravec1985high,KinectFusion, gupta2017cognitive,song2015robot}, agent localization~\cite{davison1998mobile,olson1998maximum,dellaert1999monte}, and planning~\cite{kuipers1991robot,wilcox1992robotic}.
Recent work investigates learned alternatives for exploration~\cite{Pathak2017CuriosityDrivenEB,chen2018learning,Burda2019ExplorationBR,Raileanu2020RIDERI,Parisi2021InterestingOC}.
Here, agents are often trained end-to-end with self-supervised rewards (e.g., curiosity \cite{Pathak2017CuriosityDrivenEB}) or supervised rewards (e.g., state visitation counts~\cite{Strehl2008AnAO,weihs2021visual, gadre2022continuous}).
Learning-based methods typically need less hand-tuning, but require millions of training steps and reward engineering.
We test both classical and learnable exploration strategies in the context of CoW to study their applicability to L-ZSON.

\mypara{Goal-conditioned navigation.}
Apart from open-ended exploration, many navigation tasks are goal-conditioned, where the agent needs to navigate to a specified position (i.e., point goal \cite{savva2017minos,xia2018gibson,yan2018chalet,Wijmans2019DDPPOLN,Chaplot2020LearningTE,gordon2019splitnet,Chattopadhyay2021RobustNavTB,Hahn2021NoRN}), view of the environment (i.e., image goal \cite{zhu2017target,Mezghani2021MemoryAugmentedRL,Ramakrishnan2022PONIPF}), or object category (i.e., object goal \cite{wortsman2019learning,chaplot2020object,chang2020semantic,yang2018visual,liang2021sscnav,AlHalah2022ZeroER, wani2020multion,Chattopadhyay2021RobustNavTB,procthor}).  
We consider an object goal navigation task.

\mypara{Instruction following in Navigation.}
Prior work investigates language-based navigation, where language provides step-by-step instructions for the task~\cite{huang2022visual,krantz2020beyond, ku2020room}. 
This line of work demonstrates the benefits of additional language input for robot navigation, especially for long-horizon tasks (e.g., room-to-room navigation~\cite{ku2020room}). However, providing detailed step-by-step instructions (e.g., move 3 meters south~\cite{huang2022visual}) could be challenging and time-consuming. 
In contrast, in our L-ZSON task, an algorithm gets natural language as the goal description instead of low-level instructions.
Our task is more challenging as it requires the agent to infer its own exploration and searching strategies. 

\mypara{Zero-shot object navigation (ZSON).}
Recent work studies object navigation in zero-shot settings, where agents are evaluated on object categories that they are not explicitly trained on~\cite{Khandelwal2021Simple,majumdar2022zson}.
Our task encompasses ZSON; however it also considers cases where more information---object attributes or hidden objects descriptions---is specified.
Khandelwal~\etal~\cite{Khandelwal2021Simple} train on a subset of \robo categories and evaluate on a held-out set.
In concurrent work, Majumdar \etal~\cite{majumdar2022zson} train on an image goal navigation task and evaluate on object navigation downstream by leveraging CLIP multi-modal embeddings. 
Both algorithms necessitate navigation training for millions of steps and train separate models for each simulation domain.
In contrast, CoW baselines do not necessitate any simulation training and can be deployed in multiple environments.

\section{The L-ZSON Task}
\label{sec:task}

Language-driven zero-shot object navigation (L-ZSON) involves navigating to goal objects, specified in language, without explicit training to do so.
Let $\mathcal{O}$ denote a set of natural language descriptions of target objects with potentially many attributes (e.g., ``plant", ``snake plant", ``plant under the bed", etc.).
This contrasts with definitions studied in prior object navigation~\cite{Batra2020ObjectNavRO,Deitke2020RoboTHOR} and ZSON~\cite{Khandelwal2021Simple,majumdar2022zson} tasks, which focus on high-level categories like ``plant".
Let $\mathcal{S}$ denote the set of navigation scenes.
Let $p_0$ describe the initial pose of an agent.
A navigation episode $\tau \in \mathcal{T}$ is written as a tuple $\tau = (s, o, p_0)$, $s \in \mathcal{S}, o \in \mathcal{O}$.
Each $\tau$ is a \emph{zero-shot task} as tuples of this form are not seen during training.
Starting at $p_0$, an embodied agent's goal is to find $o$.
The agent receives observations and sensor readings $I_t$ (e.g., RGB-D images).
At each timestep $t$, the agent executes a navigation action $a \in \mathcal{A}$.
A special action $\textsc{Stop} \in \mathcal{A}$ terminates the episode.
If the agent is within $c$ units of $o$ and $o$ meets a visibility criteria, the episode is successful.

\section{CLIP on Wheels (CoW) Baselines}
\label{sec:cow}

\begin{figure}[tp]
    \centering
    \includegraphics[width=\linewidth]{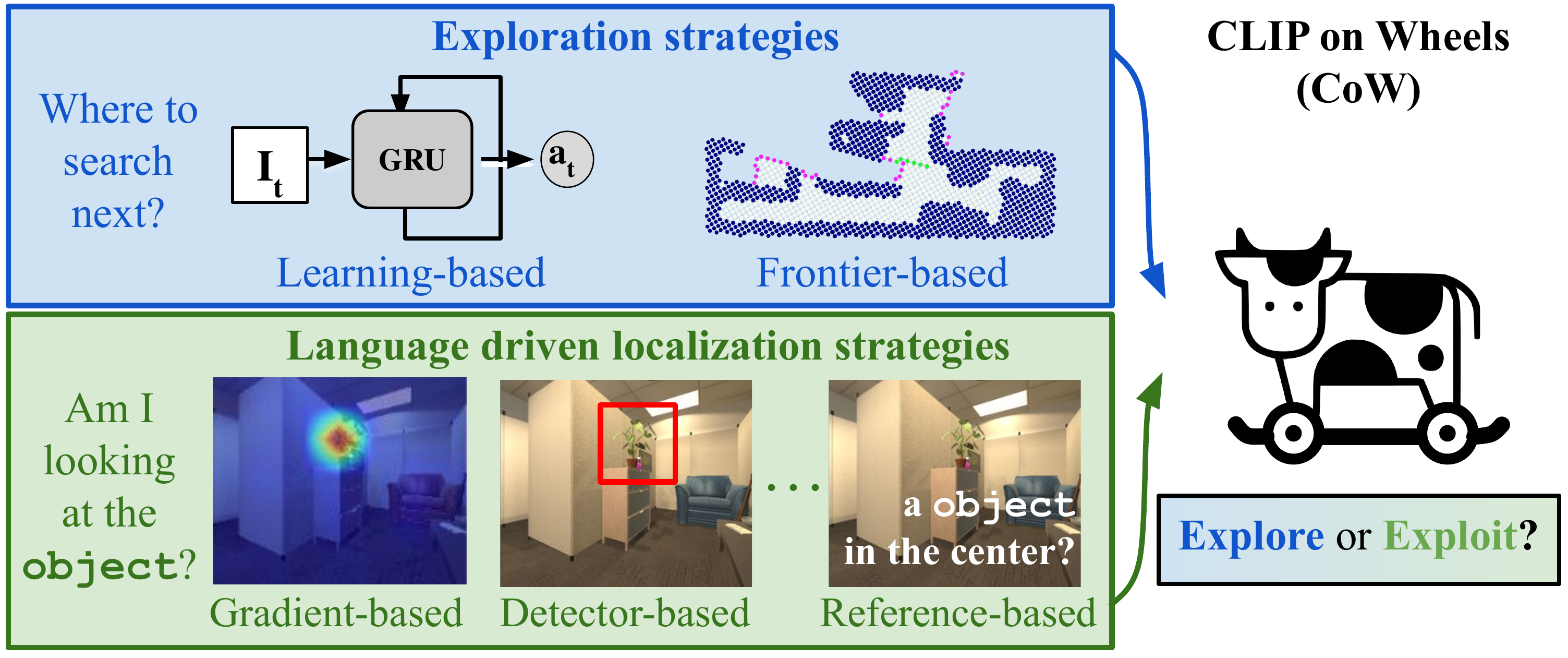}
    \vspace*{-5mm}
    \caption{\textbf{CLIP on Wheels (CoW) overview.}
    A CoW uses a policy to explore and an object localizer (e.g., an open-vocabulary object detector) to determine if an object goal is in view.
    }
    \vspace{-5mm}
    \label{fig:cow}
\end{figure}

To address L-ZSON, we investigate a simple baseline approach, CoW, which adapts open-vocabulary models like CLIP to make them suitable for the task.
A CoW takes as input an egocentric RGB-D image and an object goal specified in language.
As a CoW moves, it updates a top-down map of the world created using RGB-D observations and pose estimates (Sec.~\ref{sec:map}).
Each CoW gets an exploration policy and a zero-shot object localization module as seen in Fig.~\ref{fig:cow}.
To observe diverse views of the scene, a CoW explores using a policy (Sec.~\ref{sec:exploration}).
As the CoW roams, it keeps track of its confidence about the target object's location using an object localization module (Sec.~\ref{sec:localization}) and its top-down map.
When a CoW's confidence exceeds a threshold, it plans to the location of the goal and issues the \textsc{Stop} action.
We now describe the ingredients used to make the CoWs evaluated in our experiments (Sec.~\ref{sec:experiments}). 

\subsection{Depth-based Mapping}
\label{sec:map}
\begin{figure}[tp]
    \centering
    \includegraphics[width=\linewidth]{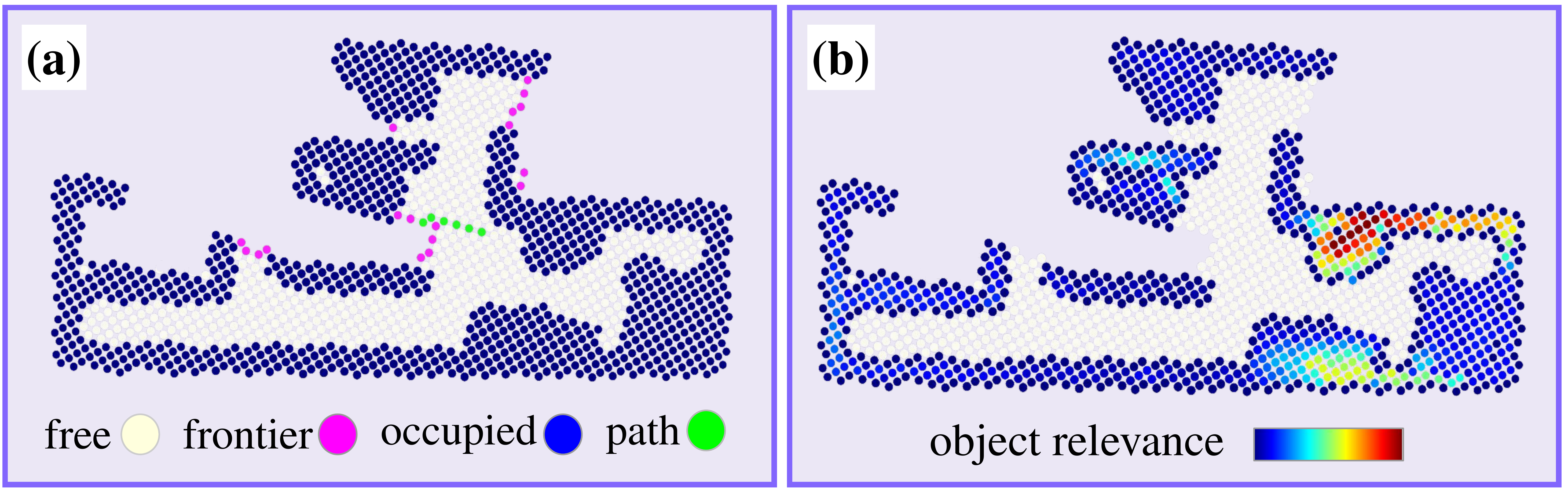}
    \vspace*{-5mm}

    \caption{\textbf{Mapping.}
    Top-down map created from egocentric depth observations as a CoW roams a space.
    (a) Frontier Based Exploration~\cite{Yamauchi1997Frontier} showing a planned path exploration path to the next frontier.
    (b) Back-projected object relevance scores provide object goal targets when a CoW has found an object.
    }
    \vspace{-5mm}
    \label{fig:map_top}
\end{figure}

As a CoW moves, it constructs a top-down map using input depth, pose estimates, and known agent height.
The map is initialized using known camera intrinsics and the first depth image.
Since a CoW knows the intended consequences of its actions (e.g., \textsc{MoveForward} should result in a 0.25m translation), each action is represented as a pose delta transform to approximate a transition.
To deal with noise associated with actuation or depth, a CoW maintains a map at 0.125m resolution.
To improve map accuracy, a CoW checks for failed actions by comparing successive depth frames for movements (see  Appx.~\ref{appx:map} for details).
Using known agent height (0.9m), map cells are projected to the ground plane to maintain a top-down representation of the world, which suffices for most navigation applications.
Cells close to the floor are considered free space (white points in Fig.~\ref{fig:map_top}~(a)), while other cells are considered occupied (blue points in Fig.~\ref{fig:map_top}~(a)).

\subsection{Exploration}
\label{sec:exploration}
Exploration generates diverse egocentric views so a CoW is more likely to view the language-specified target object.
We consider two exploration methods, \emph{frontier-based} and 
\emph{learning-based}.

\mypara{Frontier based exploration (FBE)~\cite{Yamauchi1997Frontier}.}
Using the top-down map discussed in Sec.~\ref{sec:map}, a CoW can navigate using a simple exploration heuristic: move to the frontier between free and unknown space to discover new regions.
Once the navigator reaches a frontier (visualized as purple points in Fig.~\ref{fig:map_top}~(a)), it moves greedily to the next closest frontier.
Since the map is updated at every timestep, a noisy pose estimate can contribute to inaccuracies.
For example, narrow passages may collapse in the map due to pose drift.
To circumvent such problems, we reinitialize the map when no paths exist to any frontiers in the map.

\mypara{Learnable exploration.}
In addition to FBE, we investigate learnable alternatives, which may explore more intelligently.
We consider an architecture and reward structure similar to prior work in embodied AI (e.g., \cite{weihs2021visual,Khandelwal2021Simple,gadre2022continuous}).
Specifically, we adopt a frozen CLIP backbone with a trainable GRU~\cite{Cho2014OnTP} and linear heads for the actor and critic networks.
We train agents independently in \hab~\cite{habitat19iccv} and \robo~\cite{Deitke2020RoboTHOR} simulation environments for 60M steps each, using DD-PPO~\cite{Schulman2017ProximalPO,Wijmans2019DDPPOLN} in the AllenAct~\cite{Weihs2020AllenActAF} framework.
We employ a simple count-based reward~\cite{Strehl2008AnAO}, which self-supervised exploration methods often attempt to approximate (e.g., \cite{Burda2019ExplorationBR}).
All training scenes are disjoint from downstream navigation test scenes; however, learnable exploration involves millions of steps of simulation training, which the FBE heuristic does not necessitate.
For details on reward, hyperparameters, and training, see Appx.~\ref{appx:exp}.

\subsection{Object Localization}
\label{sec:localization}

Successful navigation depends on object localization: the ability to tell \emph{if} and \emph{where} an object is in an image.
Regions of high object relevance, extracted from 2D images, are projected to the depth-based map (Fig.~\ref{fig:map_top}~(b)) where they serve as natural navigation targets.
To determine if and when a target is in an image, we consider the following object localization modules, used in our experiments (Sec.~\ref{sec:experiments}).
For more details see Appx.~\ref{appx:loc}.

\mypara{CLIP~\cite{Radford2021LearningTV} with $k$  referring expressions.}
We match a CLIP visual embedding for the current observation against $k$ different CLIP text embeddings, which specify potential locations of the target object.
For example, ``a plant in the top left of the image."
CLIP computes similarity between the image and text features to determine relevance scores over the language specified image regions.

\mypara{CLIP~\cite{Radford2021LearningTV} with $k$ image patches.}
We discretize the image into $k$ smaller patches, and run CLIP vision backbone inference on each of them to obtain patch features.
We then convolve each patch feature with a CLIP text embedding for the target object.
If the object is in a patch, the relevance score for that patch should be high.

\mypara{CLIP~\cite{Radford2021LearningTV} with gradient relevance~\cite{Chefer2021TransformerIB}.}
We adapt a network interpretability method~\cite{Selvaraju2019GradCAM,Chefer2021TransformerIB} to extract object relevancy from vision transformers (ViTs)~\cite{Dosovitskiy2021AnII}.
Using a target CLIP text embedding and gradient information accumulated through the CLIP vision backbone, Chefer~\etal~\cite{Chefer2021TransformerIB} construct a relevance map over the image pixels, which qualitatively segments the target.
These relevance methods typically assume the target object is visible and normalized between zero and one.
We observe that removing this normalization produces high relevance when the target is in view and low relevance otherwise, hence providing signal for true positive and true negative detections.

\mypara{MDETR segmentation~\cite{Kamath2021MDETRM}.}
Kamath~\etal~\cite{Kamath2021MDETRM} extend the DETR detector~\cite{Carion2020EndtoEndOD} to take arbitrary text and images as input and output box detections.
They fine-tune their base model on PhraseCut~\cite{Wu2020PhraseCutLI}, a dataset of paired masks and attribute descriptions, to support segmentation.
We use this MDETR model to obtain object relevance for targets directly.

\mypara{OWL-ViT detection~\cite{Minderer2022SimpleOO}.} Minderer~\etal~\cite{Minderer2022SimpleOO} study a recipe to turn CLIP-like models into object detectors by fine-tuning on a set prediction task.
Similar to MDETR, we use OWL-ViT to directly query images for targets.

\mypara{Post-processing.}
The aforementioned models give continuous valued predictions.
However, we are interested in binary masks giving if and where objects are in images.
Hence, we threshold predictions for each model (see Appx.~\ref{appx:loc} for details).
We further investigate two strategies for using the masks downstream: (1) using the entire mask or (2) using only the center pixel of masks.
The second strategy is reasonable because only part of an object needs to be detected for successful navigation.

\mypara{Target driven planning.}
Recall, CoWs have depth sensors.
We back-project object relevance from 2D images into the depth-based map (Sec.~\ref{sec:map}).
We keep only the max relevance for each map cell (Fig.~\ref{fig:map_top}~(b)).
CoWs can then plan to high relevance areas in the map.
See Appx.~\ref{appx:moreviz} for additional method visualization.
\section{The \dataset Benchmark}
\label{sec:pasture}

\begin{figure}[tp]
    \centering
    \includegraphics[width=\linewidth]{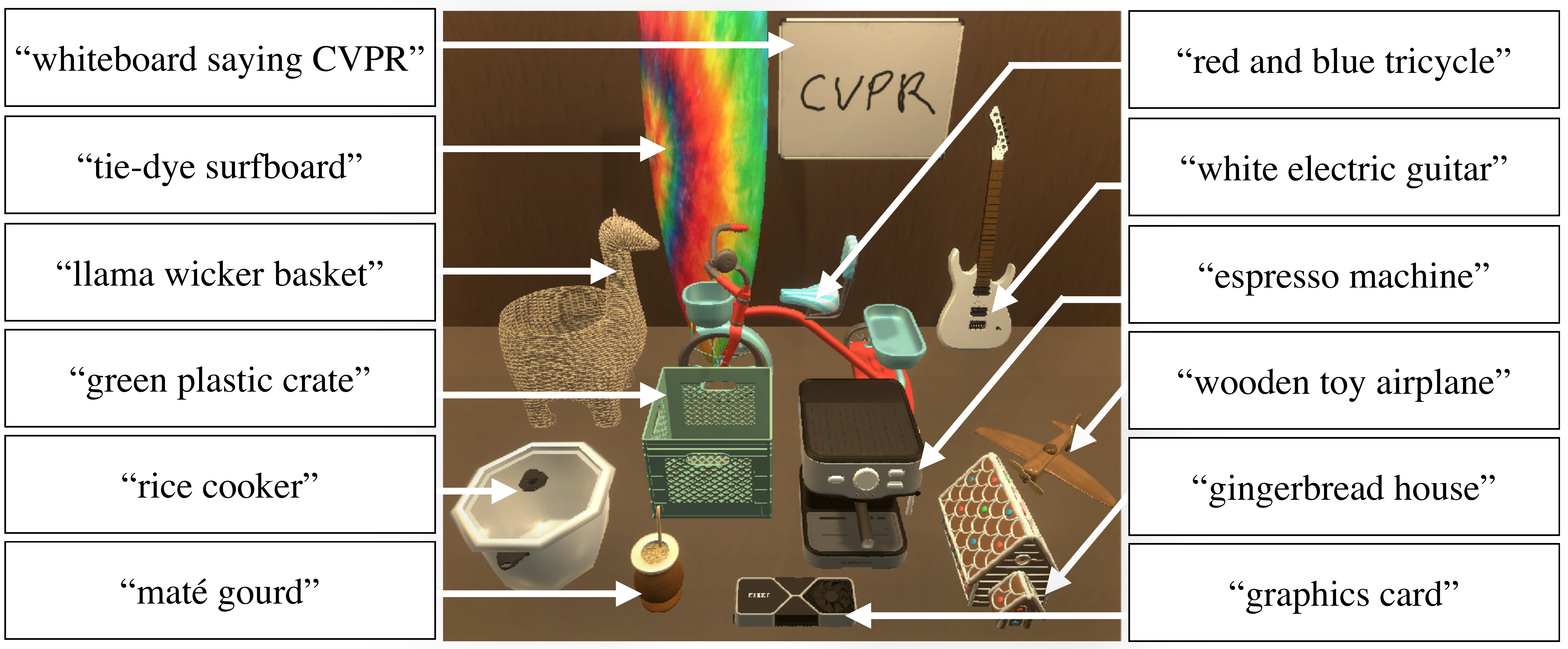}
    \vspace*{-5mm}
    \caption{\textbf{Uncommon objects} in \dataset.
    }
    \vspace{-5mm}
    \label{fig:uncommon}
\end{figure} 

To evaluate CoW baselines and future methods on L-ZSON, we introduce the \dataset evaluation benchmark.
\dataset builds on \robo validation scenes, which have parallel environments in the real-world.
We target \robo to facilitate future real-world benchmarking.
\dataset probes for seven capabilities explained in Sec.~\ref{sec:capabilities}.
We provide dataset statistics in Sec.~\ref{sec:stats}.

\subsection{\textbf{\dataset} Tasks}
\label{sec:capabilities}

\dataset evaluates seven core L-ZSON capabilities.

\mypara{Uncommon objects.} 
Traditional benchmarks (e.g., \robo and \hab MP3D) evaluate agents on common object categories like TVs; however, given the rich diversity of objects in homes, we would like to understand navigation performance on \emph{uncommon objects}.
Hence we add 12 new objects to each room.
We use names shown in Fig.~\ref{fig:uncommon} as instance labels, which are minimal descriptions to identify each object.
Some identifiers refer to text in images (e.g., ``whiteboard saying CVPR") or to appearance attributes (e.g., ``wooden toy airplane").
Other objects are less common in North America, like ``mat\'{e}", which is a popular Argentinian drink.

\mypara{Appearance descriptions.}
To evaluate if baselines can take advantage of visual attributes, we introduce descriptions of the form: ``\{\emph{size}\}, \{\emph{color}\}, \{\emph{material}\} \{\emph{object}\}".
For example: ``small, red apple", ``orange basketball", ``small, black, metallic alarm clock".
Objects are considered small if their 3D bounding box diagonal is below a threshold. We determine color and materials by inspection.

\mypara{Spatial descriptions.}
To test if agents can leverage spatial information in navigation, we introduce descriptions: ``\{\emph{object}\} on top of \{$x$\}, near \{$y$\}, \{$z$\}, ...".
For example, ``house plant on a dresser near a spray bottle".
To determine [on top of] relations, we use \textsc{THOR} metadata and to determine [nearness] we use a distance threshold between pairs of objects.
We inspect all descriptions for correctness.

\mypara{Appearance descriptions with distractors.}
To probe if appearance attributes better facilitate finding objects in the presence of distractors, we reuse the appearance captions from before, but evaluate on an modified environment with two visually distinct instances of each \robo object category.
For example, for the task of finding a ``red apple", we have both a red apple and a green apple in the room.
A navigator must leverage appearance information---and not just class information---to successfully complete the task.

\mypara{Spatial descriptions with distractors.}
This capability is similar to the one above; however, we evaluate with spatial descriptions in the presence of distractor objects.

\mypara{Hidden object descriptions.}
An ideal object navigator should find objects, even when they are hidden.
To test this capability, we introduce descriptions: ``\{\emph{object}\} under/in \{\emph{x}\}".
For example, ``basketball in the dresser drawers" or ``vase under the sofa".
We sample large objects (e.g., beds, sofas, dressers) in each scene to determine [under/in] relations.
Additionally we remove visible instances of \{\emph{object}\} from the room.

\mypara{Hidden object descriptions with distractors.}
We use the hidden object descriptions from before, but reintroduce visible instances of \{\emph{object}\} to serve as distractors.
Consider finding a ``mug under the bed".
A distractor mug will also appear in the scene making the task more challenging.

\subsection{Dataset Creation and Statistics}
\label{sec:stats}
\dataset contains three variations for each of the original 15 validation \robo rooms: uncommon objects added, additional object instances added, and target objects removed.
For each of the seven settings presented above, we evaluate over 12 object instances in 15 rooms with two initial agent starting locations.
Hence \dataset consists of 2,520 tasks, which is a similar order of magnitude to \robo (1,800) and \hab MP3D (2,195) validation sets.
For appearance attributes, 47\% of the objects are considered ``small". Each object gets an average of 1.2 color descriptors out of 22 possible choices, and 0.6 material descriptors out of 5 possible choices.
Similarly, for spatial attributes, each object gets one object it is on top of or in (e.g., ``vase in a shelving unit'') and an average of 1.9 objects it is near.
For a sample of appearance and spatial attributes see Fig.~\ref{fig:teaser}.
For more dataset details and statistics see Appx.~\ref{appx:dataset}.
\begin{table*}
\centering
\scriptsize
\begin{tabular}{cllc|ccccccc|yc|yc}
\toprule
 & & & &\multicolumn{9}{c|}{\dataset} & \multicolumn{2}{c}{\robo}\\
 \multicolumn{4}{c|}{CoW breeds} & Uncom. & Appear. & Space & Appear. & Space & Hid. & Hid. & \multicolumn{2}{c|}{Avg.} & \multicolumn{2}{c}{}\\
 & & & & & & & distract & distract & & \multicolumn{1}{c|}{distract} & & & &\\
ID & Localizer & Arch. & Post & \textsc{{SR}} & \textsc{{SR}} & \textsc{{SR}} & \textsc{{SR}} & \textsc{{SR}} & \textsc{{SR}} & \textsc{{SR}} & \textsc{{SPL}} & \textsc{{SR}} & \textsc{{SPL}} & \textsc{{SR}}\\\midrule
\textcolor{tab:pink}{\Large{$\blacktriangle$}} & CLIP-Ref. & B/32 & & 2.8& 1.4& 1.4& 0.8& 1.4& 4.7& 5.0& 1.2 & 2.5 & 1.6 & 2.2\\
\textcolor{tab:pink}{\Large{$\triangle$}} & CLIP-Ref. & B/32 & \checkmark& 3.6& 0.6& 1.7& 0.6& 1.7& 2.2& 2.5& 0.9 & 1.8 & 1.0 & 1.8\\
\textcolor{tab:pink}{\Large{$\blacksquare$}} & CLIP-Ref. & B/16 & & 1.4& 1.7& 1.7& 1.9& 1.9& 2.8& 2.2& 1.7 & 1.9 & 2.4 & 2.6\\
\textcolor{tab:pink}{\Large{$\square$}} & CLIP-Ref. & B/16 & \checkmark& 1.4& 2.8& 2.8& 3.1& 3.3& 1.7& 1.9& 1.7 & 2.4 & 2.1 & 2.7\\\midrule
\textcolor{tab:orange}{\Large{$\blacktriangle$}} & CLIP-Patch & B/32 & & 10.6& 9.7& 6.7& 6.4& 6.4& 16.7& 16.7& 7.5 & 10.4 & 9.0 & 14.3\\
\textcolor{tab:orange}{\Large{$\triangle$}} & CLIP-Patch & B/32 & \checkmark& 18.1& 13.3& 13.3& 8.6& 10.8& 17.5& \textbf{17.8}& 9.0 & 14.2 & 10.6 & 20.3\\
\textcolor{tab:orange}{\Large{$\blacksquare$}} & CLIP-Patch & B/16 & & 5.6& 7.8& 3.9& 5.0& 3.9& 10.6& 10.8& 5.4 & 6.8 & 8.2 & 10.3\\
\textcolor{tab:orange}{\Large{$\square$}} & CLIP-Patch & B/16 & \checkmark& 10.6& 11.4& 7.8& 10.8& 8.1& 16.4& 15.6& 7.7 & 11.5 & 9.7 & 15.7\\\midrule
\textcolor{tab:purple}{\Large{$\blacktriangle$}} & CLIP-Grad. & B/32 & & 13.6& 10.6& 9.2& 7.5& 7.2& 13.9& 12.8& 8.3 & 10.7 & 9.6 & 13.8\\
\textcolor{tab:purple}{\Large{$\triangle$}} & CLIP-Grad. & B/32 & \checkmark& 16.1& 11.9& 11.7& 9.7& 10.3& 14.4& 16.1& 9.2 & 12.9 & 9.7 & 15.2\\
\textcolor{tab:purple}{\Large{$\blacksquare$}} & CLIP-Grad. & B/16 & & 6.1& 5.8& 5.0& 5.0& 4.7& 8.3& 6.9& 4.9 & 6.0 & 7.3 & 8.8\\
\textcolor{tab:purple}{\Large{$\square$}} & CLIP-Grad. & B/16 & \checkmark& 8.1& 10.8& 8.6& 8.6& 6.7& 11.1& 11.4& 6.7 & 9.3 & 8.6 & 11.6\\\midrule
\textcolor{tab:green}{\Large{$\blacklozenge$}} & MDETR & B3 & & 3.1& 6.9& 4.4& 7.2& 4.7& 7.8& 8.9& 5.3 & 6.2 & 8.3 & 9.8\\
\textcolor{tab:green}{\Large{$\lozenge$}} & MDETR & B3 & \checkmark& 3.1& 7.2& 5.0& 6.9& 4.7& 8.1& 8.9& 5.4 & 6.3 & 8.4 & 9.9\\\midrule
\textcolor{tab:cyan}{\Large{$\blacktriangle$}} & OWL & B/32 & & 23.1& 26.1& 14.4& 18.3& 11.7& 13.9& 13.1& 11.1 & 17.2 & 16.6 & 25.4\\
\textcolor{tab:cyan}{\Large{$\triangle$}} & OWL & B/32 & \checkmark& \textbf{32.8}& 26.4& \textbf{19.4}& \textbf{19.4}& \textbf{16.1}& \textbf{19.2}& 14.4& \textbf{12.6} & \textbf{21.1} & 16.9 & 26.7\\
\textcolor{tab:cyan}{\Large{$\blacksquare$}} & OWL & B/16 & & 25.8& 23.6& 15.3& 17.2& 12.5& 13.1& 13.9& 11.4 & 17.3 & 16.2 & 24.8\\
\textcolor{tab:cyan}{\Large{$\square$}} & OWL & B/16 & \checkmark& 31.9& \textbf{26.9}& 18.9& \textbf{19.4}& 14.7& 18.1& 15.8& \textbf{12.6} & 20.8 & \textbf{17.2} & \textbf{27.5}\\\midrule
% \multicolumn{4}{c|}{Supervised SoTA} & & & & & & & & & & &\\
% \multicolumn{4}{c|}{} & & & & & & & & & & &\\
\multicolumn{4}{c|}{\textcolor{gray}{ProcTHOR fine-tune (supervised)}~\cite{procthor}} & \textcolor{gray}{n/a} & \textcolor{gray}{n/a} & \textcolor{gray}{n/a} & \textcolor{gray}{n/a} & \textcolor{gray}{n/a} & \textcolor{gray}{n/a} & \textcolor{gray}{n/a} & \textcolor{gray}{n/a} & \textcolor{gray}{n/a} & \textcolor{gray}{27.4} & \textcolor{gray}{66.4}\\

\bottomrule
\end{tabular}
\vspace*{-1mm}
\caption{\textbf{Benchmarking CoWs on \dataset for L-ZSON.} On \dataset we identify several key takeaways.
(1) Average success on \dataset is lower than on \robo; however, CoWs are surprisingly good at finding uncommon objects (Uncom.), often finding them at higher rates than more common \robo objects.
(2) Comparing filled ($\newmoon$) vs. unfilled ($\fullmoon$) row IDs, we notice post-processing mask predictions by using only the center pixel as a representative target for navigation helps in general (see Sec.~\ref{sec:localization} for more details on post-processing).
(3) Comparing square ($\square$) vs. triangle ($\triangle$) IDs, we see that architectures (Arch.) using \emph{more compute} (i.e., ViT-B/16) often perform comparably or worse than their competitors (i.e., ViT-B/32). This is especially true for CLIP~\cite{Radford2021LearningTV} models (indicated in \textcolor{tab:pink}{pink}, \textcolor{tab:orange}{orange}, and \textcolor{tab:purple}{purple}).
(4) \textcolor{tab:cyan}{Blue} OWL-ViT~\cite{Minderer2022SimpleOO} models perform best.
(5) \dataset tasks with distractor objects (distract) hurt performance and natural language specification is not sufficient to mitigate against the added difficulties in these tasks.
(6) A supervised baseline shown in \textcolor{gray}{gray} significantly outperforms CoWs on \robo; however, it is unable to support \dataset tasks out-of-the-box.
}
\label{'tab:main'}
\vspace*{-5mm}
\end{table*}

\section{Experiments}
\label{sec:experiments}

We first present our experimental setup, including the datasets, metrics, embodiment, and baselines considered in our study (Sec.~\ref{sec:experimental_setup}).
Then we present results on \dataset, thereby elucidating the strengths and weakness of CoW baselines for L-ZSON (Sec.~\ref{sec:pasture_results}).
Finally, we compare to prior ZSON art in \robo and \hab (MP3D) environments (Sec.~\ref{sec:e2e}).

\subsection{Experimental setup}
\label{sec:experimental_setup}

\mypara{Environments.}
We consider \dataset (Sec.~\ref{sec:pasture}), \robo~\cite{Deitke2020RoboTHOR}, and \hab (MP3D)~\cite{habitat19iccv} validation sets as our test sets.
We utilize validation sets for testing because official test set ground-truth is not publicly available.
Domains are setup with
noise that is faithful to their original challenge settings.
For \robo---and by extension \dataset---this means actuation noise but no depth noise.
For \hab this means considerable depth noise and reconstruction artifacts, but no actuation noise.

\mypara{Navigation Metrics.}
We adopt standard object navigation metrics to measure performance:
\begin{itemize}[leftmargin=3mm]
    \vspace{-2mm}
    \item \textsc{Success (SR)}: the fraction of episodes where the agent executes \textsc{Stop} within 1.0m of the target object.
    \vspace{-2mm}
    \item Success weighted by inverse path length (\textsc{SPL}): Success weighted by the oracle shortest path length and normalized by the actual path length~\cite{Batra2020ObjectNavRO}. This metric points to the success efficiency of the agent.
    \vspace{-1mm}
\end{itemize}
In \robo and \dataset, the target must additionally be visible for the episode to be a success, which this is not the case in \hab---as specified in their 2021 challenge.

\mypara{Embodiment.}
The agent is a LoCoBot~\cite{Gupta2018RobotLI}.
All agents have discrete actions: \textsc{\{MoveForward, RotateRight, RotateLeft, Stop\}}.
The move action advances the agent by 0.25m, while rotation actions pivot the camera
by 30$^\circ$.

\mypara{CoW Baselines.}
For exploration we consider policies presented in Sec.~\ref{sec:exploration}: FBE heuristic exploration, learnable exploration optimized on \hab (MP3D) train scenes, and learned exploration optimized on \robo train scenes.
Learned exploration requires training in simulation---which is counter to our zero-shot goals; nonetheless, we ablate these explorers to contextualize their performance within the CoW framework.
\emph{FBE is the default CoW exploration strategy.}

For object localization, we consider:
\begin{itemize}[leftmargin=3mm]
    \vspace{-2mm}
    \item CLIP with $k=9$ referring expressions (\textcolor{tab:pink}{CLIP-Ref.})
    \vspace{-2mm}
    \item CLIP with $k=9$ patches (\textcolor{tab:orange}{CLIP-Patch})
    \vspace{-2mm}
    \item CLIP with gradient relevance (\textcolor{tab:purple}{CLIP-Grad.})
    \vspace{-2mm}
    \item MDETR segmentation model (\textcolor{tab:green}{MDETR})
    \vspace{-2mm}
    \item OWL-ViT detector (\textcolor{tab:cyan}{OWL}) 
    \vspace{-2mm}
    
\end{itemize}
Descriptions of these models are in Sec.~\ref{sec:localization} and additional details are in Appx.~\ref{appx:loc}.
All models are open-vocabulary. \emph{No models are fine-tuned on navigation}, and hence we consider their inference \emph{zero-shot} on our tasks.\footnote{The claim is not that these models have never seen any synthetic data in their large-scale training sets, only that they are not trained to navigate.}
We also consider various backbone architectures:
\begin{itemize}[leftmargin=3mm]
    \vspace{-2mm}
    \item A vision transformer~\cite{Dosovitskiy2021AnII}, ViT-B/32 ($\blacktriangle$ B/32)
    \vspace{-2mm}
    \item ViT-B/16 ($\blacksquare$ B/16), which uses a smaller patch size of 16x16 and hence more compute.
    \vspace{-2mm}
    \item EfficentNet B3 ($\blacklozenge$ B3), which is convolutional and similar in compute requirements to a ViT/B32.
    \vspace{-2mm}
\end{itemize}
For every model we additionally evaluate with post-processing ($\triangle$, $\square$, $\lozenge$), where only the center pixel of detections is registered in the top-down CoW map.
Recall, this is a sensible strategy as only some part of the object needs to be found for an episode to be successful.
For details on hyperparameters, learned agents, object localization threshold tuning, and CLIP prompt-tuning, see Appx.~\ref{appx:loc},~\ref{appx:prompt_ablation}.

\mypara{End-to-end learnable baselines.}
We also compare against the following methods, which are trained in simulation for millions of steps:
\begin{itemize}[leftmargin=3mm]
    \item \emph{EmbCLIP-ZSON}~\cite{Khandelwal2021Simple}: trains a model on eight \robo categories, using CLIP language embeddings to specify the goal objects. At test time, the model is evaluated on four held-out object categories. The unseen target objects are specified with category names using CLIP language embeddings.
    \vspace{-2mm}
    \item \emph{SemanticNav-ZSON}~\cite{majumdar2022zson}: trains models separately---one for each dataset---for image goal navigation, where image goals are specified with CLIP visual embeddings. At test time image goals are switched with CLIP language embeddings for the object goals.
    We compare to the MP3D trained model.
    \vspace{-2mm}
\end{itemize}
Both EmbCLIP-ZSON and SemanticNav-ZSON leverage multi-modal CLIP visual and language embeddings in learnable frameworks that require simulation training.

\begin{figure*}[tp]
    \centering
    \includegraphics[width=0.98\linewidth]{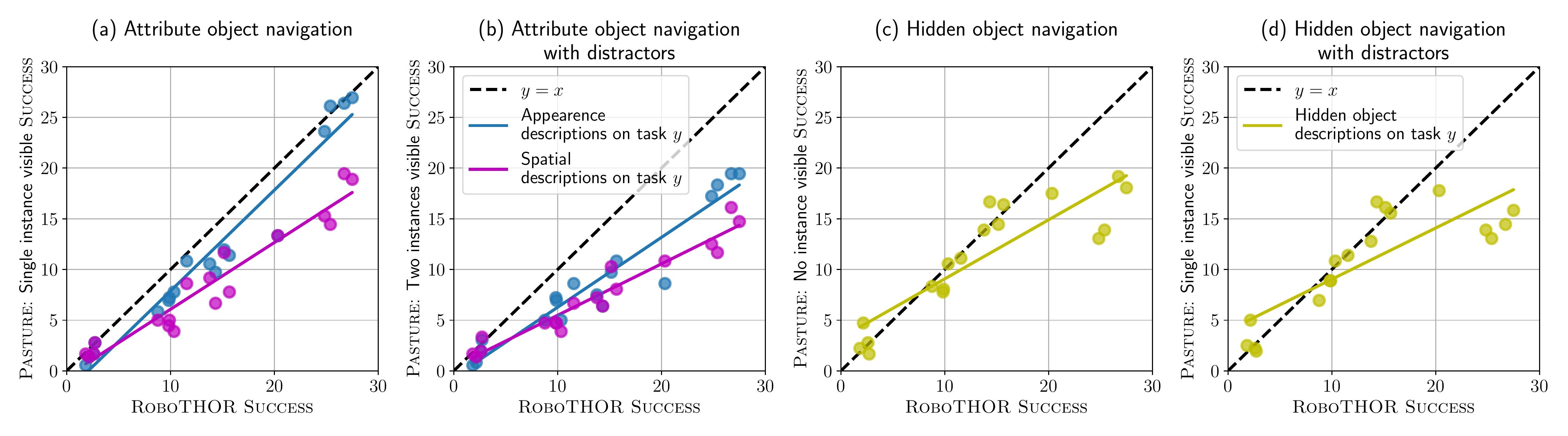} \vspace{-4mm}
    \caption{\textbf{\dataset object navigation with descriptions.}
    In general object navigation with descriptions is more challenging than the \robo object navigation task, as indicated by trend lines lying below the $y=x$ line. 
    (a) Appearance descriptions are more helpful than spatial descriptions.
    (b) Performance further drops when distractor objects are introduced to the environment. However, CoWs are still able to make better use of appearance description than spatial descriptions.
    (c) Models in the lower success regime ($<$15\% \robo \textsc{Success}) perform comparably on finding hidden objects. However, this trend plateaus for higher success models.
    (d) Trends are similar when distractor objects are introduced for hidden object navigation.
    }
    \vspace*{-3mm}
    \label{fig:pasture_splits}
\end{figure*}

\subsection{CoWs on \textbf{\dataset}}
\label{sec:pasture_results}

Tab.~\ref{'tab:main'} shows the results of 18 CoWs evaluated on \dataset. 
For category-level results see Appx.~\ref{appx:cat}.
We now discuss several salient questions.

\mypara{How well can CoWs find common objects vs. uncommon objects?}
Comparing \robo and uncommon (Uncom.) \dataset success rate (\textsc{SR}) in Tab.~\ref{'tab:main'}---first and last columns---we notice that CoWs often find uncommon objects at higher rates than common \robo objects (e.g., by $\sim$6 percentage points \textsc{Success} for the OWL ViT-B/32 CoW with post-post processing (\textcolor{tab:cyan}{$\triangle$})).
We hypothesize that though uncommon objects are less prevalent in daily life, they are still represented in open-vocabulary datasets and hence recognizable for the object localization modules.
We further explore this hypothesis in Appx.~\ref{appx:dataset} by visualizing CLIP retrieval results on LAION-5B~\cite{schuhmann2022laion} for the uncommon object categories.
The relatively high performance on uncommon objects speaks to the flexibility of CoW baselines and their ability to inherit desirable properties from the open-vocabulary models.

\mypara{Can CoWs utilize appearance and spatial descriptions?}
Looking at Fig.~\ref{fig:pasture_splits}~(a) we see that neither appearance nor spatial descriptions improve CoW performance compared to their \robo baseline performance (i.e., most points lie under the $y = x$ line).
However, CoW is able to take better advantage of appearance descriptions than spatial descriptions.
These results motivate future investigation on open-vocabulary object localization with a greater focus on textual object attributes. 

\mypara{Can CoWs find visible objects in the presence of distractors?}
In Fig.~\ref{fig:pasture_splits}~(b) we see that CoWs experience further performance degradation when compared to Fig.~\ref{fig:pasture_splits}~(a).
We conclude that appearance and spatial attributes specified as natural language input are not sufficient to deal with the added complexity of distractors given current open-vocabulary models.

\mypara{Can CoWs find hidden objects?}
Looking at Fig.~\ref{fig:pasture_splits}~(c) we notice that models in the lower success regime (less that 15\% \textsc{Success} on \robo) are able to find hidden objects at about the same rate as \robo objects (i.e., they lie on the $y = x$ line).
OWL models in the higher success regime ($>$15\%) do not continue this trend; however, they do achieve higher absolute accuracy as seen in Tab.~\ref{'tab:main'}.
Dealing with occlusion is a longstanding problem in computer vision, and these results provide a foundation upon which future hidden object navigation work can improve.

\mypara{Can CoWs find hidden objects in the presence of distractors?}
Comparing Figs.~\ref{fig:pasture_splits}~(c) and~(d), we notice similar trends lines, with the best models performing worse with distractors.
This suggests that distractors are less of a concern in the case of hidden objects than for visible object targets.
In light of the fact that detection methods generally work better on larger objects, we hypothesize this effect is because distractor objects are smaller (e.g., apples, vases, basketballs) than objects used to conceal target categories (e.g., beds, sofas, etc.).

\mypara{What exploration method performs best?}
\begin{table}[tp]
\centering
\tabcolsep=0.05cm
\scriptsize
\begin{tabular}{cccc|l|yc|yc}
\toprule
\multicolumn{5}{c|}{CoW breeds} & \multicolumn{2}{c|}{\dataset (Avg.)} &\multicolumn{2}{c}{\robo}\\
ID & Loc. & Arch. & Post  & Exp. Strategy  & \textsc{SPL} & \textsc{SR} & \textsc{SPL} & \textsc{SR}\\\midrule
\textcolor{tab:cyan}{\Large{$\triangle$}} & OWL & B/32 & \checkmark  & \robo learn. & 10.2 & 17.3 & 13.1 & 20.9\\
\textcolor{tab:cyan}{\Large{$\triangle$}} & OWL & B/32 & \checkmark  & \hab learn. & 8.6 & 19.4 & 9.8 & 20.4\\
\textcolor{tab:cyan}{\Large{$\triangle$}} & OWL & B/32 & \checkmark & FBE & \textbf{12.6} & \textbf{21.1} & \textbf{16.9} & \textbf{26.7}\\\bottomrule

\end{tabular}
\vspace*{-1mm}
\caption{\textbf{Exploration ablation.} For a fixed object localizer (OWL-ViT B/32 with post processing), we ablate over different choices of exploration policy: the FBE heuristic, agents trained in \robo, and \hab (MP3D). We find that FBE outperforms learnable alternatives on both \dataset and \robo. \hab learnable model perform worst, but are not trained on any \dataset or \robo data.
}
\label{tab:learn}
\vspace{-3mm}
\end{table}
We ablate the decision to use FBE for most experiments by fixing an object localizer (OWL, B/32, with post-processing (\textcolor{tab:cyan}{$
\triangle$})) and comparing against \robo learnable exploration and \hab learnable exploration in Tab.~\ref{tab:learn}.
We notice that FBE performs best in all cases; however, learnable exploration still performs well suggesting that these models do learn useful strategies for the downstream tasks.
Furthermore, the worse performance of the \hab learned model suggests that training on a certain domain can hurt in cases of inference in other domains (i.e., \hab~$\longrightarrow$~\robo or \hab~$\longrightarrow$~\dataset).
Future work might re-consider learning-based algorithms for navigation that perform well out-of-distribution (i.e., ``sim2real" and ``sim2sim" exploration transfer as in~\cite{gordon2019splitnet}).

\begin{figure}[tp]
    \centering
    \includegraphics[width=0.8\linewidth]{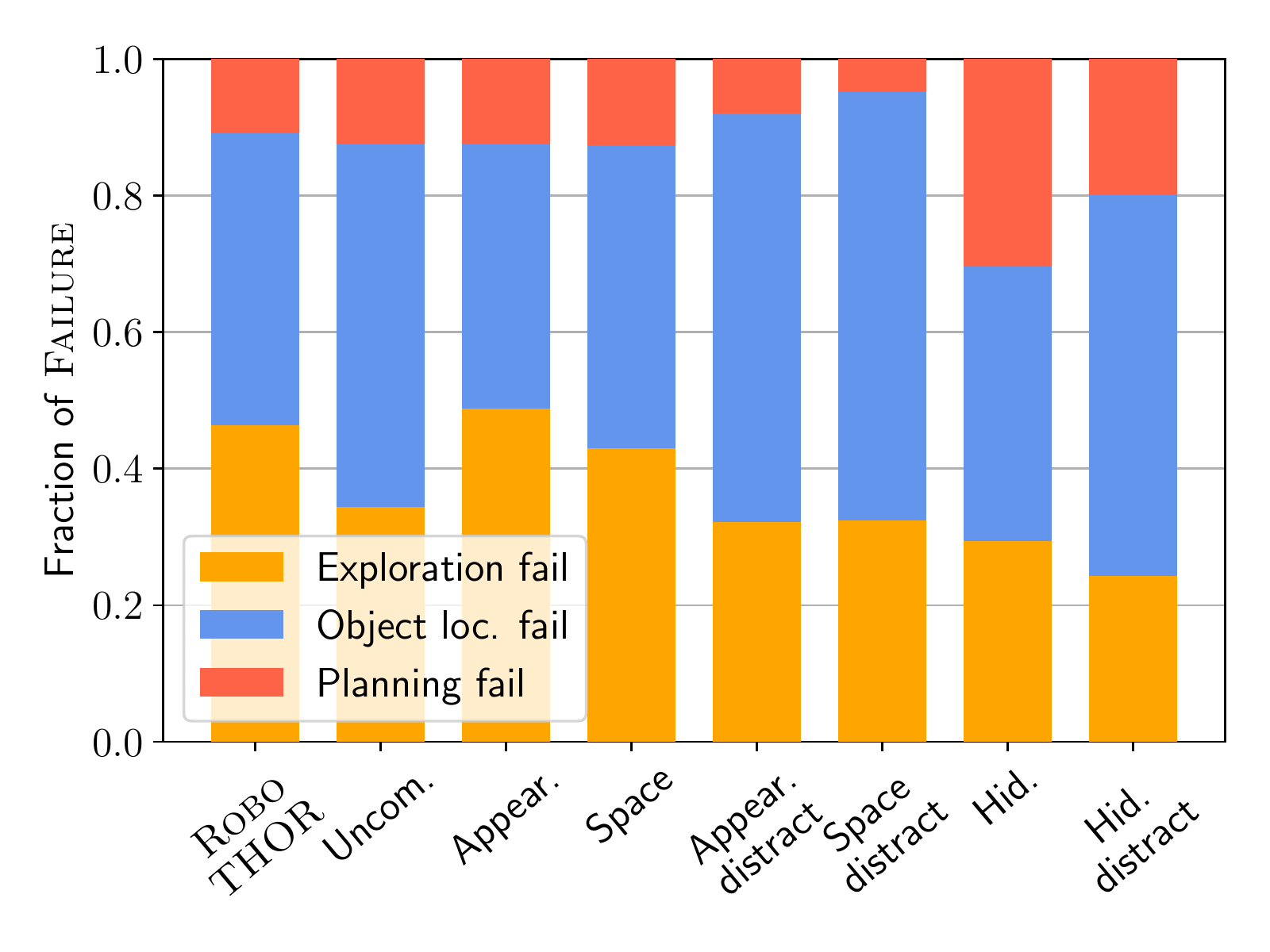}
    \vspace*{-2mm}
    \caption{\textbf{Failure analysis for OWL, B/32, post-processing (\textcolor{tab:cyan}{$\triangle$}).}
    Exploration and object localization errors occur at similar ratios, with increased localization failures in the presense of distractors.
    }
    \vspace*{-2mm}
    \label{fig:fail}
\end{figure}

\mypara{How do CoWs fail?} We identify three high-level failure modes. (1) \emph{Exploration fail}: the target is never seen. (2) \emph{Object localization fail}: the target is seen but the localizer never fires. (3) \emph{Planning fail}: the target is seen and the localizer fires, but planning fails due to inaccuracy in the map representation (Sec.~\ref{sec:exploration}).
Looking at Fig.~\ref{fig:fail}, we notice a large fraction of failures are due to exploration and object localization. This suggests CoWs may continue to improve as research in these fields progress.
In Fig.~\ref{fig:fail} we additionally see that in cases where distractors are present a higher fraction of object localization failures occurs, supporting the claim that open-vocabulary models currently struggle to make full use of attribute prompts.
See Appx.~\ref{appx:fail} for additional failure analysis.

\subsection{Comparison to Prior Art}
\label{sec:e2e}

While our primary aim is to evaluate CoWs in general L-ZSON settings, we further evaluate CoWs in ZSON settings that prior work considers to establish CoW as a strong baseline for these tasks.
Recall, ZSON can be seen as a special case of L-ZSON where only object goals are specified in language (no attributes).

In Tab.~\ref{tab:e2e}, we see there exists a CoW that outperforms the end-to-end baselines in all cases except SemanticNav-ZSON \textsc{Success} on \hab (MP3D).
For instance, the CLIP-Grad., B/32, with post-processing (\textcolor{tab:purple}{$\triangle$}) matches the SemanticNav-ZSON model on \hab (MP3D) \textsc{SPL}---4.9 for CoW v.s. 4.8 for the competitor, while also improving over EmbCLIP-ZSON \robo by 15.6 percentage points.
To contextualize the significance of this result, we reiterate that this CoW is trained for \textbf{0} navigation steps, while SemanticNav-ZSON and EmbCLIP-ZSON train in the target evaluation simulators for 500M and 60M steps respectively.

The superior performance of SemanticNav-ZSON in terms of MP3D \textsc{Success} indicates that there can be benefits to in-domain learning over CoW baselines.
Future work may consider unifying the benefits of CoW-like models and fine-tuned models to get the best of both worlds.

\begin{table}[tp]
\centering
\tabcolsep=0.07cm
\scriptsize
\begin{tabular}{cccc|yc|yc|yc|a}
\toprule
\multicolumn{4}{c|}{} & \multicolumn{2}{c|}{\hab}  & \multicolumn{2}{c|}{\robo}  & \multicolumn{2}{c|}{\robo} & Nav.\\
\multicolumn{4}{c|}{CoW breeds} &\multicolumn{2}{c|}{(MP3D)}  & \multicolumn{2}{c|}{(subset)}  & \multicolumn{2}{c|}{(full)} & training\\
ID & Loc. & Arch. & Post  & \textsc{SPL} & \textsc{SR} & \textsc{SPL} & \textsc{SR} & \textsc{SPL} & \textsc{SR} & steps\\\midrule
\textcolor{tab:purple}{\Large{$\triangle$}} & CLIP-Grad. & B/32 & \checkmark  & \textbf{4.9} & 9.2 & 15.0  & 23.7 & 9.7 & 15.2 & \textbf{0}\\
\textcolor{tab:cyan}{\Large{$\triangle$}} & OWL & B/32 & \checkmark  & 3.7 & 7.4 & \textbf{20.8}  & \textbf{32.5} & \textbf{16.9} & \textbf{26.7} & \textbf{0}\\\midrule
\multicolumn{4}{c|}{EmbCLIP-ZSON~\cite{Khandelwal2021Simple}} & --  & -- & --  & 8.1 & -- & 14.0$^*$ & 60M\\\midrule
\multicolumn{4}{c|}{SemanticNav-ZSON~\cite{majumdar2022zson}}& 4.8 & \textbf{15.3} & --  & -- & -- & -- & 500M\\

\bottomrule
\end{tabular}
\caption{\textbf{Comparison to prior art on existing ZSON benchmarks.}
CoWs are able to match or out-compete existing methods that leverage millions of steps of navigation training in the evaluation simulator.
$^*$indicates a result from prior work that includes, non-zero-shot evaluation. Specifically, 1/4 of the evaluations are zero-shot on \robo (subset) and the remaining 3/4 on categories seen during training.
} 
\vspace*{-3mm}
\label{tab:e2e}
\end{table}
\section{Limitations and Conclusion}

\mypara{Limitations.}
While our evaluation of CoWs on \hab, \robo, and \dataset is a step towards assessing their performance in different domains, ultimately, real-world performance matters most.
Hence, the biggest limitation of our work is the lack of large-scale, real-world benchmarking---which is also missing in much of the related literature.
Additionally, CoW inherents the meta-limitations of the object localization and exploration methods considered.
For example, object localizers require tuning a confidence threshold to balance precision and recall.
Finally, we do not consider different agent embodiment or continuous action spaces.
This is a pertinent investigation given recent findings of Pratt \etal~\cite{Pratt2022TheIA} that agent morphology can be a big determinant of downstream performance.

\mypara{Conclusion.} This paper introduces the \dataset benchmark for language-driven zero-shot object navigation and several CLIP on Wheels baselines, translating the successes of existing zero-shot models to an embodied task.
We view CoW as an instance of using open-vocabulary models, with text-based interfaces, to tackle robotics tasks in more flexible settings.
We hope that the baselines and the proposed benchmark will spur the field to explore broader and more powerful forms of zero-shot embodied AI. 

%%%%%%%%% REFERENCES
\clearpage

{\small
\mypara{Acknowledgement.}
We would like to thank Jessie Chapman, Cheng Chi, Huy Ha, Zeyi Liu, Sachit Menon, and Sarah Pratt for valuable feedback.
This work was supported in part by NSF CMMI-2037101, NSF IIS-2132519, and an Amazon Research Award.
SYG is supported by a NSF Graduate Research Fellowship.
The views and conclusions contained herein are those of the authors and should not be
interpreted as necessarily representing the official policies, either expressed or implied, of the sponsors.

\bibliographystyle{ieee_fullname}
\bibliography{egbib}

\begin{thebibliography}{10}\itemsep=-1pt

\bibitem{AlHalah2022ZeroER}
Ziad Al-Halah, Santhosh~K. Ramakrishnan, and Kristen Grauman.
\newblock Zero experience required: Plug \& play modular transfer learning for
  semantic visual navigation.
\newblock {\em arXiv}, 2022.

\bibitem{Baek_2021_ICCV}
Donghyeon Baek, Youngmin Oh, and Bumsub Ham.
\newblock Exploiting a joint embedding space for generalized zero-shot semantic
  segmentation.
\newblock {\em ICCV}, 2021.

\bibitem{bansal2018zero}
Ankan Bansal, Karan Sikka, Gaurav Sharma, Rama Chellappa, and Ajay Divakaran.
\newblock Zero-shot object detection.
\newblock {\em ECCV}, 2018.

\bibitem{Batra2020ObjectNavRO}
Dhruv Batra, Aaron Gokaslan, Aniruddha Kembhavi, O. Maksymets, R. Mottaghi, M.
  Savva, A. Toshev, and Erik Wijmans.
\newblock Objectnav revisited: On evaluation of embodied agents navigating to
  objects.
\newblock {\em arXiv}, abs/2006.13171, 2020.

\bibitem{NEURIPS2019_0266e33d}
Maxime Bucher, Tuan-Hung VU, Matthieu Cord, and Patrick P\'{e}rez.
\newblock Zero-shot semantic segmentation.
\newblock {\em NeurIPS}, 2019.

\bibitem{Burda2019ExplorationBR}
Yuri Burda, Harrison Edwards, Amos~J. Storkey, and Oleg Klimov.
\newblock Exploration by random network distillation.
\newblock {\em ICLR}, 2018.

\bibitem{Carion2020EndtoEndOD}
Nicolas Carion, Francisco Massa, Gabriel Synnaeve, Nicolas Usunier, Alexander
  Kirillov, and Sergey Zagoruyko.
\newblock End-to-end object detection with transformers.
\newblock {\em ECCV}, 2020.

\bibitem{chang2020semantic}
Matthew Chang, Arjun Gupta, and Saurabh Gupta.
\newblock Semantic visual navigation by watching youtube videos.
\newblock {\em NeurIPS}, 2020.

\bibitem{chaplot2020object}
Devendra~Singh Chaplot, Dhiraj Gandhi, Abhinav Gupta, and Ruslan Salakhutdinov.
\newblock Object goal navigation using goal-oriented semantic exploration.
\newblock {\em NeurIPS}, 2020.

\bibitem{Chaplot2020LearningTE}
Devendra~Singh Chaplot, Dhiraj Gandhi, Saurabh Gupta, Abhinav~Kumar Gupta, and
  Ruslan Salakhutdinov.
\newblock Learning to explore using active neural slam.
\newblock {\em ICLR}, 2020.

\bibitem{Chattopadhyay2021RobustNavTB}
Prithvijit Chattopadhyay, Judy Hoffman, Roozbeh Mottaghi, and Aniruddha
  Kembhavi.
\newblock Robustnav: Towards benchmarking robustness in embodied navigation.
\newblock {\em ICCV}, 2021.

\bibitem{Chefer2021TransformerIB}
Hila Chefer, Shir Gur, and Lior Wolf.
\newblock Transformer interpretability beyond attention visualization.
\newblock {\em CVPR}, 2021.

\bibitem{chen2018learning}
Tao Chen, Saurabh Gupta, and Abhinav Gupta.
\newblock Learning exploration policies for navigation.
\newblock {\em ICLR}, 2019.

\bibitem{Cheng_2021_ICCV}
Jiaxin Cheng, Soumyaroop Nandi, Prem Natarajan, and Wael Abd-Almageed.
\newblock Sign: Spatial-information incorporated generative network for
  generalized zero-shot semantic segmentation.
\newblock {\em ICCV}, 2021.

\bibitem{Cho2014OnTP}
Kyunghyun Cho, Bart van Merrienboer, Dzmitry Bahdanau, and Yoshua Bengio.
\newblock On the properties of neural machine translation: Encoder–decoder
  approaches.
\newblock In {\em SSST@EMNLP}, 2014.

\bibitem{davison1998mobile}
Andrew~J Davison and David~W Murray.
\newblock Mobile robot localisation using active vision.
\newblock {\em ECCV}, 1998.

\bibitem{Deitke2020RoboTHOR}
Matt Deitke, Winson Han, Alvaro Herrasti, Aniruddha Kembhavi, Eric Kolve,
  Roozbeh Mottaghi, Jordi Salvador, Dustin Schwenk, Eli VanderBilt, Matthew
  Wallingford, Luca Weihs, Mark Yatskar, and Ali Farhadi.
\newblock Robothor: An open simulation-to-real embodied ai platform.
\newblock {\em CVPR}, 2020.

\bibitem{procthor}
Matt Deitke, Eli VanderBilt, Alvaro Herrasti, Luca Weihs, Jordi Salvador, Kiana
  Ehsani, Winson Han, Eric Kolve, Ali Farhadi, Aniruddha Kembhavi, and Roozbeh
  Mottaghi.
\newblock Procthor: Large-scale embodied ai using procedural generation.
\newblock {\em NeurIPS}, 2022.

\bibitem{dellaert1999monte}
Frank Dellaert, Dieter Fox, Wolfram Burgard, and Sebastian Thrun.
\newblock Monte carlo localization for mobile robots.
\newblock {\em ICRA}, 1999.

\bibitem{Demirel2018ZeroShotOD}
Berkan Demirel, Ramazan~Gokberk Cinbis, and Nazli Ikizler-Cinbis.
\newblock Zero-shot object detection by hybrid region embedding.
\newblock {\em BMVC}, 2018.

\bibitem{Deng2009ImageNetAL}
Jia Deng, Wei Dong, Richard Socher, Li-Jia Li, K. Li, and Li Fei-Fei.
\newblock Imagenet: A large-scale hierarchical image database.
\newblock {\em CVPR}, 2009.

\bibitem{dosovitskiy2020image}
Alexey Dosovitskiy, Lucas Beyer, Alexander Kolesnikov, Dirk Weissenborn,
  Xiaohua Zhai, Thomas Unterthiner, Mostafa Dehghani, Matthias Minderer, Georg
  Heigold, Sylvain Gelly, et~al.
\newblock An image is worth 16x16 words: Transformers for image recognition at
  scale.
\newblock {\em arXiv}, 2020.

\bibitem{Dosovitskiy2021AnII}
Alexey Dosovitskiy, Lucas Beyer, Alexander Kolesnikov, Dirk Weissenborn,
  Xiaohua Zhai, Thomas Unterthiner, Mostafa Dehghani, Matthias Minderer, Georg
  Heigold, Sylvain Gelly, Jakob Uszkoreit, and Neil Houlsby.
\newblock An image is worth 16x16 words: Transformers for image recognition at
  scale.
\newblock {\em ICLR}, 2020.

\bibitem{gadre2022continuous}
Samir~Yitzhak Gadre, Kiana Ehsani, Shuran Song, and Roozbeh Mottaghi.
\newblock Continuous scene representations for embodied ai.
\newblock {\em CVPR}, 2022.

\bibitem{gordon2019splitnet}
Daniel Gordon, Abhishek Kadian, Devi Parikh, Judy Hoffman, and Dhruv Batra.
\newblock Splitnet: Sim2sim and task2task transfer for embodied visual
  navigation.
\newblock {\em CVPR}, 2019.

\bibitem{Gu2021ZeroShotDV}
Xiuye Gu, Tsung-Yi Lin, Weicheng Kuo, and Yin Cui.
\newblock Zero-shot detection via vision and language knowledge distillation.
\newblock {\em arXiv}, 2021.

\bibitem{Gupta2019LVISAD}
Agrim Gupta, Piotr Doll{\'a}r, and Ross~B. Girshick.
\newblock Lvis: A dataset for large vocabulary instance segmentation.
\newblock {\em CVPR}, 2019.

\bibitem{Gupta2018RobotLI}
Abhinav~Kumar Gupta, Adithyavairavan Murali, Dhiraj Gandhi, and Lerrel Pinto.
\newblock Robot learning in homes: Improving generalization and reducing
  dataset bias.
\newblock {\em NeurIPS}, 2018.

\bibitem{gupta2017cognitive}
Saurabh Gupta, James Davidson, Sergey Levine, Rahul Sukthankar, and Jitendra
  Malik.
\newblock Cognitive mapping and planning for visual navigation.
\newblock {\em CVPR}, pages 2616--2625, 2017.

\bibitem{Hahn2021NoRN}
Meera Hahn, Devendra~Singh Chaplot, and Shubham Tulsiani.
\newblock No rl, no simulation: Learning to navigate without navigating.
\newblock {\em NeurIPS}, 2021.

\bibitem{henry2014rgb}
Peter Henry, Michael Krainin, Evan Herbst, Xiaofeng Ren, and Dieter Fox.
\newblock Rgb-d mapping: Using depth cameras for dense 3d modeling of indoor
  environments.
\newblock {\em Experimental robotics}, 2014.

\bibitem{NEURIPS2020_f73b76ce}
Ping Hu, Stan Sclaroff, and Kate Saenko.
\newblock Uncertainty-aware learning for zero-shot semantic segmentation.
\newblock {\em NeurIPS}, 2020.

\bibitem{huang2022visual}
Chenguang Huang, Oier Mees, Andy Zeng, and Wolfram Burgard.
\newblock Visual language maps for robot navigation.
\newblock {\em arXiv preprint arXiv:2210.05714}, 2022.

\bibitem{jia2021scaling}
Chao Jia, Yinfei Yang, Ye Xia, Yi-Ting Chen, Zarana Parekh, Hieu Pham, Quoc~V
  Le, Yunhsuan Sung, Zhen Li, and Tom Duerig.
\newblock Scaling up visual and vision-language representation learning with
  noisy text supervision.
\newblock {\em ICML}, 2021.

\bibitem{Kamath2021MDETRM}
Aishwarya Kamath, Mannat Singh, Yann LeCun, Ishan Misra, Gabriel Synnaeve, and
  Nicolas Carion.
\newblock Mdetr - modulated detection for end-to-end multi-modal understanding.
\newblock {\em ICCV}, 2021.

\bibitem{Kato_2019_ICCV}
Naoki Kato, Toshihiko Yamasaki, and Kiyoharu Aizawa.
\newblock Zero-shot semantic segmentation via variational mapping.
\newblock {\em ICCV-W}, 2019.

\bibitem{Khandelwal2021Simple}
Apoorv Khandelwal, Luca Weihs, Roozbeh Mottaghi, and Aniruddha Kembhavi.
\newblock Simple but effective: Clip embeddings for embodied ai.
\newblock {\em arXiv}, 2021.

\bibitem{krantz2020beyond}
Jacob Krantz, Erik Wijmans, Arjun Majumdar, Dhruv Batra, and Stefan Lee.
\newblock Beyond the nav-graph: Vision-and-language navigation in continuous
  environments.
\newblock In {\em European Conference on Computer Vision}, pages 104--120.
  Springer, 2020.

\bibitem{ku2020room}
Alexander Ku, Peter Anderson, Roma Patel, Eugene Ie, and Jason Baldridge.
\newblock Room-across-room: Multilingual vision-and-language navigation with
  dense spatiotemporal grounding.
\newblock {\em arXiv preprint arXiv:2010.07954}, 2020.

\bibitem{kuipers1991robot}
Benjamin Kuipers and Yung-Tai Byun.
\newblock A robot exploration and mapping strategy based on a semantic
  hierarchy of spatial representations.
\newblock {\em Robotics and autonomous systems}, 1991.

\bibitem{Li2019ZeroShotOD}
Zhihui Li, Lina Yao, Xiaoqin Zhang, Xianzhi Wang, Salil~S. Kanhere, and
  Huaxiang Zhang.
\newblock Zero-shot object detection with textual descriptions.
\newblock {\em AAAI}, 2019.

\bibitem{liang2021sscnav}
Yiqing Liang, Boyuan Chen, and Shuran Song.
\newblock Sscnav: Confidence-aware semantic scene completion for visual
  semantic navigation.
\newblock {\em ICRA}, 2021.

\bibitem{lin_2015}
Tsung-Yi Lin, Michael Maire, Serge Belongie, Lubomir Bourdev, Ross Girshick,
  James Hays, Pietro Perona, Deva Ramanan, C.~Lawrence Zitnick, and Piotr
  Dollár.
\newblock Microsoft coco: Common objects in context.
\newblock {\em ECCV}, 2014.

\bibitem{majumdar2022zson}
Arjun Majumdar, Gunjan Aggarwal, Bhavika Devnani, Judy Hoffman, and Dhruv
  Batra.
\newblock Zson: Zero-shot object-goal navigation using multimodal goal
  embeddings.
\newblock {\em arXiv preprint arXiv:2206.12403}, 2022.

\bibitem{Mao2020ZeroShotOD}
Qiaomei Mao, Chong Wang, Sheng Yu, Ye Zheng, and Yuqi Li.
\newblock Zero-shot object detection with attributes-based category similarity.
\newblock {\em IEEE Transactions on Circuits and Systems II: Express Briefs},
  2020.

\bibitem{Mezghani2021MemoryAugmentedRL}
Lina Mezghani, Sainbayar Sukhbaatar, Thibaut Lavril, Oleksandr Maksymets, Dhruv
  Batra, Piotr Bojanowski, and Alahari Karteek.
\newblock Memory-augmented reinforcement learning for image-goal navigation.
\newblock {\em arXiv}, 2021.

\bibitem{Minderer2022SimpleOO}
Matthias Minderer, Alexey~A. Gritsenko, Austin Stone, Maxim Neumann, Dirk
  Weissenborn, Alexey Dosovitskiy, Aravindh Mahendran, Anurag Arnab, Mostafa
  Dehghani, Zhuoran Shen, Xiao Wang, Xiaohua Zhai, Thomas Kipf, and Neil
  Houlsby.
\newblock Simple open-vocabulary object detection with vision transformers.
\newblock {\em ECCV}, 2022.

\bibitem{montavon2017explaining}
Gr{\'e}goire Montavon, Sebastian Lapuschkin, Alexander Binder, Wojciech Samek,
  and Klaus-Robert M{\"u}ller.
\newblock Explaining nonlinear classification decisions with deep taylor
  decomposition.
\newblock {\em Pattern recognition}, 65:211--222, 2017.

\bibitem{moravec1985high}
Hans Moravec and Alberto Elfes.
\newblock High resolution maps from wide angle sonar.
\newblock {\em ICRA}, 1985.

\bibitem{KinectFusion}
Richard~A Newcombe, Shahram Izadi, Otmar Hilliges, David Molyneaux, David Kim,
  Andrew~J Davison, Pushmeet Kohi, Jamie Shotton, Steve Hodges, and Andrew
  Fitzgibbon.
\newblock Kinectfusion: Real-time dense surface mapping and tracking.
\newblock {\em ISMAR}, 2011.

\bibitem{Nicholson2019QuadricSLAMDQ}
Lachlan Nicholson, Michael Milford, and Niko S{\"u}nderhauf.
\newblock Quadricslam: Dual quadrics from object detections as landmarks in
  object-oriented slam.
\newblock {\em RA-L}, 2019.

\bibitem{olson1998maximum}
Clark~F Olson and Larry~H Matthies.
\newblock Maximum likelihood rover localization by matching range maps.
\newblock {\em ICRA}, 1998.

\bibitem{Parisi2021InterestingOC}
Simone Parisi, Victoria Dean, Deepak Pathak, and Abhinav~Kumar Gupta.
\newblock Interesting object, curious agent: Learning task-agnostic
  exploration.
\newblock {\em NeurIPS}, 2021.

\bibitem{Pathak2017CuriosityDrivenEB}
Deepak Pathak, Pulkit Agrawal, Alexei~A. Efros, and Trevor Darrell.
\newblock Curiosity-driven exploration by self-supervised prediction.
\newblock {\em ICML}, 2017.

\bibitem{pham2021scaling}
Hieu Pham, Zihang Dai, Golnaz Ghiasi, Hanxiao Liu, Adams~Wei Yu, Minh-Thang
  Luong, Mingxing Tan, and Quoc~V. Le.
\newblock Combined scaling for zero-shot transfer learning.
\newblock {\em arXiv}, 2021.

\bibitem{Pratt2022TheIA}
Sarah Pratt, Luca Weihs, and Ali Farhadi.
\newblock The introspective agent: Interdependence of strategy, physiology, and
  sensing for embodied agents.
\newblock {\em arXiv}, 2022.

\bibitem{Radford2021LearningTV}
Alec Radford, Jong~Wook Kim, Chris Hallacy, Aditya Ramesh, Gabriel Goh,
  Sandhini Agarwal, Girish Sastry, Amanda Askell, Pamela Mishkin, Jack Clark,
  Gretchen Krueger, and Ilya Sutskever.
\newblock Learning transferable visual models from natural language
  supervision.
\newblock {\em ICML}, 2021.

\bibitem{Rahman2018ZeroShotOD}
Shafin Rahman, Salman~Hameed Khan, and Fatih~Murat Porikli.
\newblock Zero-shot object detection: Learning to simultaneously recognize and
  localize novel concepts.
\newblock {\em ACCV}, 2018.

\bibitem{Raileanu2020RIDERI}
Roberta Raileanu and Tim Rockt{\"a}schel.
\newblock Ride: Rewarding impact-driven exploration for procedurally-generated
  environments.
\newblock {\em ICLR}, 2020.

\bibitem{Ramakrishnan2022PONIPF}
Santhosh~K. Ramakrishnan, Devendra~Singh Chaplot, Ziad Al-Halah, Jitendra
  Malik, and Kristen Grauman.
\newblock Poni: Potential functions for objectgoal navigation with
  interaction-free learning.
\newblock {\em CVPR}, 2022.

\bibitem{russakovsky_2015}
Olga Russakovsky, Jia Deng, Hao Su, Jonathan Krause, Sanjeev Satheesh, Sean Ma,
  Zhiheng Huang, Andrej Karpathy, Aditya Khosla, Michael Bernstein,
  Alexander~C. Berg, and Li Fei-Fei.
\newblock Imagenet large scale visual recognition challenge.
\newblock {\em IJCV}, 2015.

\bibitem{savva2017minos}
Manolis Savva, Angel~X Chang, Alexey Dosovitskiy, Thomas Funkhouser, and
  Vladlen Koltun.
\newblock Minos: Multimodal indoor simulator for navigation in complex
  environments.
\newblock {\em arXiv}, 2017.

\bibitem{habitat19iccv}
Manolis Savva, Abhishek Kadian, Oleksandr Maksymets, Yili Zhao, Erik Wijmans,
  Bhavana Jain, Julian Straub, Jia Liu, Vladlen Koltun, Jitendra Malik, Devi
  Parikh, and Dhruv Batra.
\newblock Habitat: {A} {P}latform for {E}mbodied {AI} {R}esearch.
\newblock {\em ICCV}, 2019.

\bibitem{schuhmann2022laion}
Christoph Schuhmann, Romain Beaumont, Richard Vencu, Cade Gordon, Ross
  Wightman, Mehdi Cherti, Theo Coombes, Aarush Katta, Clayton Mullis, Mitchell
  Wortsman, et~al.
\newblock Laion-5b: An open large-scale dataset for training next generation
  image-text models.
\newblock {\em NeurIPS}, 2022.

\bibitem{Schulman2017ProximalPO}
John Schulman, Filip Wolski, Prafulla Dhariwal, Alec Radford, and Oleg Klimov.
\newblock Proximal policy optimization algorithms.
\newblock {\em arXiv}, 2017.

\bibitem{Selvaraju2019GradCAM}
Ramprasaath~R. Selvaraju, Abhishek Das, Ramakrishna Vedantam, Michael Cogswell,
  Devi Parikh, and Dhruv Batra.
\newblock Grad-cam: Visual explanations from deep networks via gradient-based
  localization.
\newblock {\em IJCV}, 2019.

\bibitem{song2015robot}
Shuran Song, Linguang Zhang, and Jianxiong Xiao.
\newblock Robot in a room: Toward perfect object recognition in closed
  environments.
\newblock {\em arXiv}, 2015.

\bibitem{Strehl2008AnAO}
Alexander~L. Strehl and Michael~L. Littman.
\newblock An analysis of model-based interval estimation for markov decision
  processes.
\newblock {\em J. Comput. Syst. Sci.}, 2008.

\bibitem{wani2020multion}
Saim Wani, Shivansh Patel, Unnat Jain, Angel~X. Chang, and Manolis Savva.
\newblock Multion: Benchmarking semantic map memory using multi-object
  navigation.
\newblock {\em NeurIPS}, 2020.

\bibitem{weihs2021visual}
Luca Weihs, Matt Deitke, Aniruddha Kembhavi, and Roozbeh Mottaghi.
\newblock Visual room rearrangement.
\newblock {\em CVPR}, 2021.

\bibitem{Weihs2020AllenActAF}
Luca Weihs, Jordi Salvador, Klemen Kotar, Unnat Jain, Kuo-Hao Zeng, Roozbeh
  Mottaghi, and Aniruddha Kembhavi.
\newblock Allenact: A framework for embodied ai research.
\newblock {\em arXiv}, 2020.

\bibitem{Wijmans2019DDPPOLN}
Erik Wijmans, Abhishek Kadian, Ari~S. Morcos, Stefan Lee, Irfan Essa, D.
  Parikh, Manolis Savva, and Dhruv Batra.
\newblock Dd-ppo: Learning near-perfect pointgoal navigators from 2.5 billion
  frames.
\newblock {\em ICLR}, 2019.

\bibitem{wilcox1992robotic}
Brian~H Wilcox.
\newblock Robotic vehicles for planetary exploration.
\newblock {\em Applied Intelligence}, 1992.

\bibitem{wortsman2019learning}
Mitchell Wortsman, Kiana Ehsani, Mohammad Rastegari, Ali Farhadi, and Roozbeh
  Mottaghi.
\newblock Learning to learn how to learn: Self-adaptive visual navigation using
  meta-learning.
\newblock {\em CVPR}, 2019.

\bibitem{Wu2020PhraseCutLI}
Chenyun Wu, Zhe Lin, Scott~D. Cohen, Trung Bui, and Subhransu Maji.
\newblock Phrasecut: Language-based image segmentation in the wild.
\newblock {\em CVPR}, 2020.

\bibitem{xia2018gibson}
Fei Xia, Amir~R Zamir, Zhiyang He, Alexander Sax, Jitendra Malik, and Silvio
  Savarese.
\newblock Gibson env: Real-world perception for embodied agents.
\newblock {\em CVPR}, 2018.

\bibitem{Yamauchi1997Frontier}
Brian Yamauchi.
\newblock A frontier-based approach for autonomous exploration.
\newblock {\em CIRA}, 1997.

\bibitem{yan2018chalet}
Claudia Yan, Dipendra Misra, Andrew Bennnett, Aaron Walsman, Yonatan Bisk, and
  Yoav Artzi.
\newblock Chalet: Cornell house agent learning environment.
\newblock {\em arXiv}, 2018.

\bibitem{yang2018visual}
Wei Yang, Xiaolong Wang, Ali Farhadi, Abhinav Gupta, and Roozbeh Mottaghi.
\newblock Visual semantic navigation using scene priors.
\newblock {\em arXiv}, 2018.

\bibitem{zhai2022lit}
Xiaohua Zhai, Xiao Wang, Basil Mustafa, Andreas Steiner, Daniel Keysers,
  Alexander Kolesnikov, and Lucas Beyer.
\newblock Lit: Zero-shot transfer with locked-image text tuning.
\newblock {\em CVPR}, 2022.

\bibitem{Zhou2021DenseCLIPEF}
Chong Zhou, Chen~Change Loy, and Bo Dai.
\newblock Denseclip: Extract free dense labels from clip.
\newblock {\em arXiv}, 2021.

\bibitem{Zhu2020ZeroSD}
Pengkai Zhu, Hanxiao Wang, and Venkatesh Saligrama.
\newblock Zero shot detection.
\newblock {\em IEEE Transactions on Circuits and Systems for Video Technology},
  2020.

\bibitem{zhu2017target}
Yuke Zhu, Roozbeh Mottaghi, Eric Kolve, Joseph~J Lim, Abhinav Gupta, Li
  Fei-Fei, and Ali Farhadi.
\newblock Target-driven visual navigation in indoor scenes using deep
  reinforcement learning.
\newblock {\em ICRA}, 2017.

\end{thebibliography}
}

\clearpage
\appendix
\tableofcontents

\section{Depth-based Mapping Details}
\label{appx:map}

\mypara{Action failure checking.}
For both learnable and heuristic explorers, actions may fail.
For map based explorers, an obstacle might be below the field of view and hence not captured as occupied space.
For learnable explorers, the output policy may heavily favor a failed action (e.g., \textsc{MoveAhead}), which could be executed repeatedly.

To improve the action success of our CoWs, we employ a simple strategy.
We compute the absolute difference between depth channels in the observations $\Delta_{i, i+1} = |D_i - D_{i+1}|$.
We then compute the mean $\mu(\Delta_{i, i+1})$ and standard deviation $\sigma(\Delta_{i, i+1})$ as representative statistics.
These quantities have interpretable meaning in meters.
Since actions should move the agent forward by approximately a fixed distance or rotation, we can then set reasonable thresholds for $\mu$, $\sigma$ below which we can be confident actions failed.
In our studies, we set these to $\mu = 0.1$m, $\sigma = 0.1$m.

Note, that in modern robot navigation systems, on-board pose estimation and bumper sensors for obstacle avoidance are nearly ubiquitous.
Hence, it is possible to implement action failure checking beyond visual inputs, even though in this paper \emph{we only consider vision-based failure checking}.

\mypara{Formalization of registering new depth observations.}
Recall, a CoW constructs a top-down map based on ego-centric depth observation and approximated poses deltas.
We provide a formalization of this process.

To create this map, we first estimate the pose of the CoW's coordinate frame.
Let $C_i$ be the current local coordinate frame of a CoW at timestep $i$.
Let $W$ denote a world frame.
We would like to align observations from each local frame to $W$ to keep a single, consistent map.
When a CoW is initialized, we set ${^{W}}\hat{T}_{C_{0}} = I$, where $I$ is the identity SE(3) transform and  ${^{W}}\hat{T}_{C_{0}}$ is a transform, which aligns frame $C_0$ to $W$.
Since a CoW knows the intended consequences of its actions (e.g., \textsc{MoveForward} should result in a translation of 0.25m), each action can be represented as a delta transform,
which models an action transition.
By concatenating these transforms over time, we get pose estimates.
To estimate pose at timestep $t+1$, we compute $^{W}\hat{T}_{C_{t+1}} = {^{W}}\hat{T}_{C_{t}} \boldsymbol{\cdot} ^{C_t}\hat{T}_{C_{t+1}}$.
Due to sensor-noise and action failures, these estimates are sensitive to pose drift.
We use the current estimated pose ${^W}T_{C_{t}}$, camera intrinsic matrix $K$, current depth observation $D_t$, standard back-projection of $D_t$, and pose concatenation to register new observations in frame $W$.

Since navigation is mostly concerned with free v.s. occupied space, we do not keep the whole 3D map.
Rather we project the map on the ground-plane using the known agent height and down gravity direction.
Points already near the floor are considered free space, while other points are considered occupied as shown in.
Additionally, we discretize the map, which additionally helps deal with sensor noise.

\section{Exploration Details}
\label{appx:exp}

\mypara{Learnable exploration.}
We use the AllenAct~\cite{Weihs2020AllenActAF} framework to conduct exploration agent training.
Our DD-PPO hyperparameters are in Tab.~\ref{tab:ddppo_hparam}.
We employ a simple state visitation count based reward.
An agent receives a reward of 0.1 for visiting a previously unvisited voxel location (at 0.125m resolution), and a step penalty of -0.01. We train two agents, one on \robo and the other on \hab MP3D train sets, which are disjoint from the downstream validation sets we use for testing.
\begin{table}
\centering
\tabcolsep=1.0cm
\begin{tabular}{l|l}
\toprule
        Hyperparameter  & Value\\\midrule
        Discount factor ($\gamma$) & 0.99\\
        GAE parameter ($\lambda$) & 0.95\\
        Clipping parameter ($\epsilon$) & 0.1\\
        Value loss coefficient & 0.5\\
        Entropy loss coefficient & 0.01\\
        LR & 3e-4\\
        Optimizer & Adam\\
        Training steps & 60M\\
        Steps per rollout & 500\\
\bottomrule
\end{tabular}
\caption{
\textbf{DD-PPO hyperparameters.}
Used for training exploration agents in both \hab and \robo.
}
\label{tab:ddppo_hparam}
\end{table}

\section{Localization Details}
\label{appx:loc}

Here we provide a review of the CLIP model~\cite{Radford2021LearningTV}, the concept of prompt ensembling as it relates to CLIP, the gradient saliency method introduced by Chefer \etal~\cite{Chefer2021TransformerIB}, MDETR~\cite{Kamath2021MDETRM}, and OWL-ViT~\cite{Minderer2022SimpleOO}.

\mypara{CLIP overview.}
Recent open-vocabulary models for zero-shot image classification include CLIP~\cite{Radford2021LearningTV},
ALIGN~\cite{jia2021scaling}, and BASIC~\cite{pham2021scaling}.
These methods pre-train contrastively on image-caption pairs from the internet.
In the case of the original CLIP model, on 400M pairs.
The key insight is that the image representation---extracted by a vision tower---and the caption representation---extracted by a text tower---should be similar.
Hence, the contrastive objects encourages image and text representations for a positive pairs to be similar and for these representations to be dissimilar from other images and captions.

More formally, CLIP jointly trains two encoders: a visual encoder $f_\theta$ and a language encoder $f_\lambda$.
Given a dataset of image-text pairs $\mathcal{D} = \{..., (I_k, t_k), (I_{k+1}, t_{k+1}), ...\}$, which can be generated at massive scale by leveraging internet data.
CLIP employs standard mini-batch style training.
Given a batch of size $n$, with $n$ image-text pairs, the current functions $f_\theta$ and $f_\lambda$ are used to featurize the data, yielding \emph{L-2 normalized} embeddings for the batch: $\{(z^I_0, z^t_0), (z^I_1, z^t_1), ... (z^I_n, z^t_n),\}$.
It is then possible to define a symmetric contrastive loss of the following form, for each sample in the batch:
\begin{equation}
    \mathcal{L}(z^I_i, z^t_i) = \frac{1}{2} \left( \frac{z^I_i \cdot z^t_i}{\sum_{j\neq i}z^I_i \cdot z^t_j} + 
    \frac{z^I_i \cdot z^t_i}{\sum_{j\neq i}z^I_j \cdot z^t_i}
    \right).
\end{equation}
By minimizing this loss, the representation for images and their corresponding captions are being pushed closer together, while these representations are being pushed farther from other images and text features in the batch, which are assumed to contain dissimilar concepts.

After training, these models can be thought to match images and captions.
Given a set of captions identifying concepts (e.g., ``a photo of an apple.", ``a photo of an orange", etc.) it is possible to construct a classification head by extracting text features, via $f_\lambda$, and L-2 normalizing each of them.
Now given an image, say of an apple, we can extract a visual feature, via $f_\theta$, and again L-2 normalize.
By dotting the visual feature with the text features, we can find the highest similarity score amongst the text captions and assign the image the corresponding label (e.g., "apple" from the caption ``a photo of an apple.").
Because this downstream classifier has not been trained to specifically classify images (say apples), it is considered a zero-shot classifier.

\mypara{CLIP prompt ensembling.}
Radford~\etal~\cite{Radford2021LearningTV} find a simple strategy to boost performance of a CLIP zero-shot classifier. Instead of using one prompt for a class (e.g., ``a photo of a apple."), they instead compute many text features for a class (e.g., ``a photo of a apple.", ``a blurry photo of a apple.", etc.). By simply averaging all these text features, they find performance improves.
For CLIP-Ref., CLIP-Patch, and CLIP-Grad. strategies we similarly use prompt ensembling using the set of 80 prompt templates used for ImageNet-1k evaluation.\footnote{\href{https://github.com/openai/CLIP/blob/e184f608c5d5e58165682f7c332c3a8b4c1545f2/notebooks/Prompt_Engineering_for_ImageNet.ipynb}{notebook exploring the effects of the 80 prompts on ImageNet-1k}}
We present a prompt ablation in Appx.~\ref{appx:prompt_ablation}.

\mypara{Grad-CAM~\cite{Selvaraju2019GradCAM} and Chefer \etal~\cite{Chefer2021TransformerIB} overview.}

We begin by reviewing the Grad-CAM formulation, from which many gradient-based interpretability methods draw inspiration.
Given a target class $c$, an input image $x$ and a model $f_\theta$, Grad-CAM produces a localization map $L_c$, capturing the importance of each pixel for the image to be classified as the target class. For standard convolutional neural networks, neuron importance weights $\alpha_k^c$ are obtained from average pooling the gradients of the activations $A^k$:

\begin{equation}
    \alpha_k^c = \textsf{AveragePool}_{i,j} \left(\frac{\partial y^c}{\partial A_{ij}^k}\right).
    \label{eq:gradcam_alpha}
\end{equation}

The relevance map is then given by a linear combination of the activations, weighted by the importance from Eq.~\ref{eq:gradcam_alpha}:

\begin{equation}
    L_c = \textsf{ReLU} \left(\sum_{k} \alpha_k^cA^k\right),
\end{equation}
\noindent where ReLU denotes the rectified linear unit function, $\textsf{ReLU}(x) = \max(0, x)$.
The final relevance map is obtained by resizing $L_c$ to the same size as the original image, using bilinear interpolation.

For ViTs \cite{dosovitskiy2020image}, we follow the method of Chefer \etal~\cite{Chefer2021TransformerIB}. Specifically, given the attention maps $A^k$ for each transformer block $k$, and relevance scores $R^k$ \cite{montavon2017explaining}, the relevance map $L_c^{\textrm{ViT}}$ is given by:

\begin{equation}
    L_c^{\textrm{ViT}} = \prod_k \bar{A}^k,
\end{equation}
\begin{equation}
    \bar{A}^k = I + \mathbb{E}_h\left[\textsf{ReLU}\left(\nabla A^k \odot R^k\right)\right],
\end{equation}
\noindent where $\mathbb{E}_h$ computes the mean over the attention heads and $\odot$ represents the element-wise product. In our case, we only look at attention maps from the final transformer block. Hence, $k=1$.

\begin{figure}[tp]
    \centering
    \includegraphics[width=\linewidth]{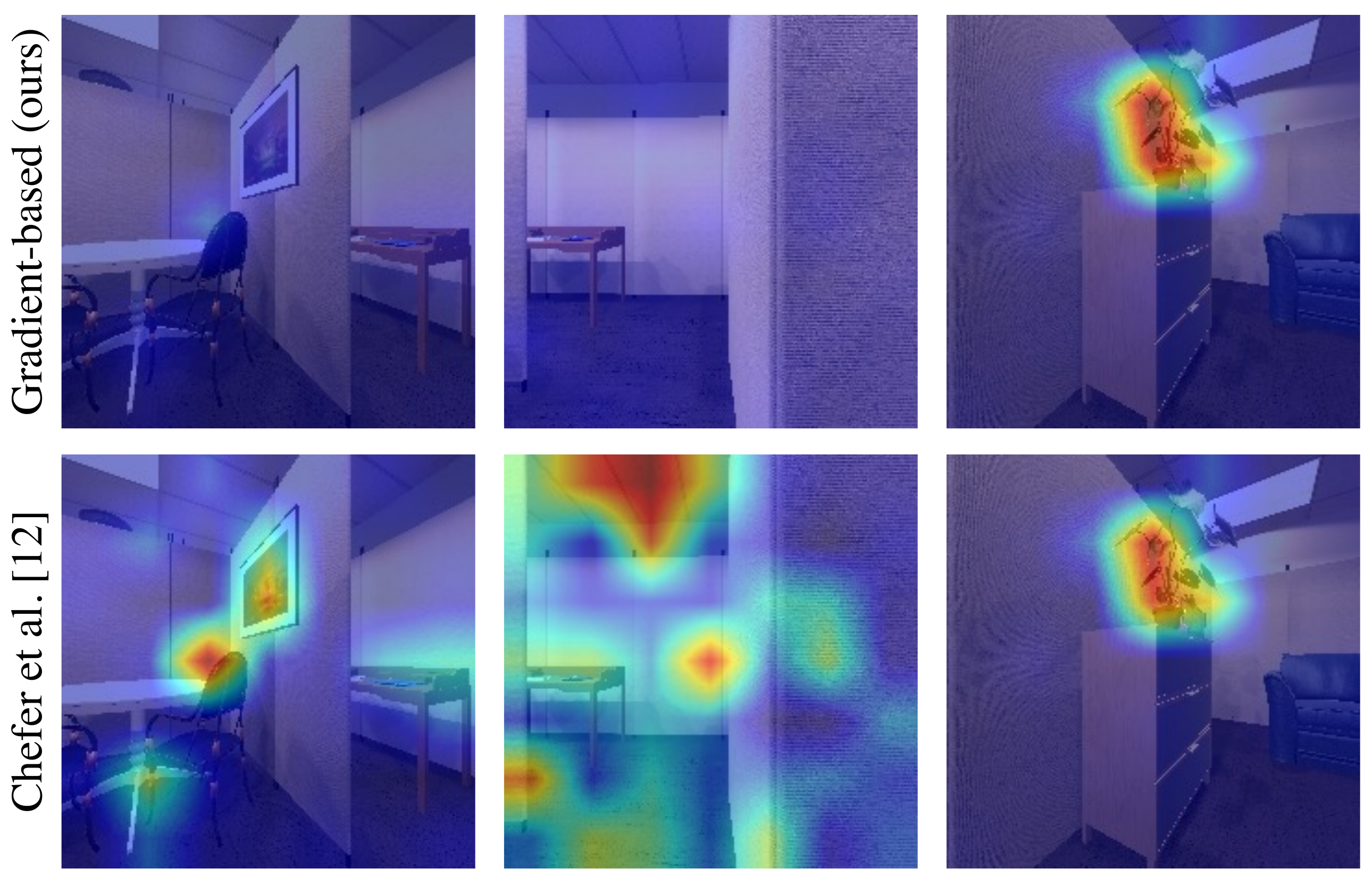}
    \caption{
        \textbf{Gradient-based relevance visualization.}
        The target object is a plant.
        For our gradient-based strategy, derived from that of Chefer \etal~\cite{Chefer2021TransformerIB}, relevance is low when the object is not in the image and high otherwise.
        In contrast the original method, produces spurious relevance when the target object is not in the frame due to the normalization it employs. Notice when the plant is in the frame, the relevance map is similar.
    }
    \label{fig:pos_negative}
    \vspace{-3mm}
\end{figure} 

It is standard to use an interpretability method like those of Selvaraju \etal~\cite{Selvaraju2019GradCAM} or Chefer \etal~\cite{Chefer2021TransformerIB} to query for a class that is known to be in the image a priori. 
Hence, it is common practice to normalize relevance maps by subtracting the minimum relevance and normalizing by the difference between the maximum and minimum relevance.
This results in a max value of 1.0.
As mentioned in Sec.~\ref{sec:localization} this is not a suitable strategy for object navigation because in many frames the object is \emph{not} in the image.
Hence, in early experiments, we removed the normalization.
Fig.~\ref{fig:pos_negative} makes the differences between these strategies apparent.
We notice this simple modification gives signal not only for true positive detections, but---critically---also for true negatives.
Notice that when the plant is not in view, relevance is qualitatively low.
However, when the plant is in view, relevance spikes in the region of the plant. 

\mypara{MDETR overview.}
MDETR~\cite{Kamath2021MDETRM} utilizes an encoder-decoder scheme to associate tokens in an input prompt referring to parts of an input image, with output boxes.
MDETR utilizes a vision tower and a pre-trained language tower to project an image, text pair into a joint embedding space (similar to CLIP as discussed above).
Image and text features are concatenated and passed to a transformer decoder head, which outputs boxes.The model is trained on a dataset of 200,000 images annotated with captions, boxes, and correspondence between words and boxes.

The model employs three losses during training. 
(1) DETR bipartite matching loss without class information: Following DETR~\cite{Carion2020EndtoEndOD}, each ground truth box is matched with its most similar predicted box and an L1 loss is applied. Unlike DETR, MDETR does not use any class information for the matching step. 
(2) A soft-token classification loss: when classifying a box, the target is a uniform distribution over all the tokens the box refers to and zero for other tokens. In this way the box is assigned to potentially many tokens depending on the ground truth.
(3) A contrastive loss in the bottleneck embedding space: this is analogous to the CLIP loss discussed above.

To fine-tune MDETR for segmentation, the original box model is fine-tuned in two stages on the PhraseCut dataset~\cite{Wu2020PhraseCutLI}.
First the model is fine-tuned for box prediction discussed above on PhraseCut data, which contains referring expressions and corresponding regions in the image (masks and boxes).
The weights are frozen and a new segmentation head is trained using Dice/F1 loss and focal loss.

\mypara{OWL ViT overview.}
OWL ViT~\cite{Minderer2022SimpleOO} employs a two-stage training procedure to create an open-vocabulary model.
During the first stage they train a CLIP-like model.
However, while the original CLIP model uses the feature corresponding to a ViT [CLS] token to construct its multi-modal embedding space, OWL ViT instead pools over patch tokens to obtain an image representation.
Intuitively, this encourages the image global feature to encode more local information from each patch.
During the second stage of training, the pooling layer is removed. The patch tokens are passed to a linear projection head where they are then dotted against text features to determine class probabilities.
An MLP box-projection head is introduced that predicts a box for each projected patch token.
Note this process and losses for box fine-tuning are similar to that used by DETR~\cite{Carion2020EndtoEndOD}.
Stage 1 is trained using 3.6 billion image-text pairs from the dataset used in LiT~\cite{zhai2022lit}, using a standard CLIP loss. Stage 2 is fine-tuned using an agglomeration of existing box datasets with $\sim$2M images total.

\mypara{Localization thresholds.}
Each object localization method discussed in Sec.~\ref{sec:localization}, requires a confidence threshold, which is standard when using a detector.
To tune this threshold, we render 500 images in \robo and 500 images in \hab MP3D with box annotations.
Critically, we use the training rooms for these datasets, so none of the scenes overlap with those seen during downstream navigation testing.
See Tab.~\ref{tab:loc_hparam} for thresholds, which are chosen to maximize an F1-score.
Because \dataset is a test set, we do \emph{no} hyper-tuning on \dataset.
Instead we use the hyper-parameters from \robo for \dataset evaluations.
\begin{table}
\centering
\scriptsize
\begin{tabular}{cll|cc}
\toprule
 & & & \hab & \robo and\\
IDs & Localizer & Arch. & & \dataset\\\midrule
\textcolor{tab:pink}{\Large{$\blacktriangle$}},\textcolor{tab:pink}{\Large{$\triangle$}} & CLIP-Ref. & B/32 & -- & 0.25\\
\textcolor{tab:pink}{\Large{$\blacksquare$}}, \textcolor{tab:pink}{\Large{$\square$}} & CLIP-Ref. & B/16 & -- & 0.125\\\midrule
\textcolor{tab:orange}{\Large{$\blacktriangle$}}, \textcolor{tab:orange}{\Large{$\triangle$}} & CLIP-Patch & B/32 & -- & 0.875\\
\textcolor{tab:orange}{\Large{$\blacksquare$}}, \textcolor{tab:orange}{\Large{$\square$}} & CLIP-Patch & B/16 & --& 0.75\\\midrule
\textcolor{tab:purple}{\Large{$\blacktriangle$}}, \textcolor{tab:purple}{\Large{$\triangle$}} & CLIP-Grad. & B/32 &  0.375 & 0.625\\
\textcolor{tab:purple}{\Large{$\blacksquare$}}, \textcolor{tab:purple}{\Large{$\square$}} & CLIP-Grad. & B/16 & -- & 0.375\\\midrule
\textcolor{tab:green}{\Large{$\blacklozenge$}}, \textcolor{tab:green}{\Large{$\lozenge$}} & MDETR & B3 & -- & 0.95\\\midrule
\textcolor{tab:cyan}{\Large{$\blacktriangle$}}, \textcolor{tab:cyan}{\Large{$\triangle$}} & OWL & B/32 & 0.2 & 0.125\\
\textcolor{tab:cyan}{\Large{$\blacksquare$}}, \textcolor{tab:cyan}{\Large{$\square$}} & OWL & B/16 & -- & 0.125\\

\bottomrule
\end{tabular}
\vspace*{-1mm}
\caption{\textbf{Object localization hyperparameters.}
Hyperparameters returned from a grid-search on object localization performance on \hab and \robo train scenes. Note, no hyperparameter tuning was conducted on \dataset, which is strictly a test set. Missing entries indicate that we did not evaluate these models due to lacking performance on other datasets (i.e., on \robo).
}
\label{tab:loc_hparam}
\vspace*{-2mm}
\end{table}

Here we present more information on computing our F1-scores.
For each image, there are some object categories $\mathcal{O}^+$ that appear in that image and other categories $\mathcal{O}^-$ that do not.
We consider a predicted binary localization mask $M^{o^+}_\text{pred}$ a true positive (TP) if $\textsf{score}^+~=~\sum(M^{o^+}_\text{pred}~\odot~M^{o^+}_\text{GT})/\sum M^{o^+}_\text{pred}~> 0.5$, where $o^+ \in \mathcal{O}^+$, $M^{o^+}_\text{GT}$ is the ground truth mask, the summation is over all binary pixel values, and $\odot$ is an element-wise product that gives binary mask intersection.
This is a more lenient measure than traditional jaccard index; however, also more applicable for our navigation setting where only part of the object needs to be identified correctly to provide a suitable navigation target.
False positives (FP) arise when $0 < \textsf{score}^+ \leq 0.5$ or $\sum M^{o^-}_\text{pred} > 0$, where $o^- \in \mathcal{O}^-$.
False negatives (FN) arise when $\sum M^{o^+} = 0$.
F1 is computed in the standard way: ${\text{TP}} / {(\text{TP} + 0.5(\text{FP} + \text{FN}))}$.
In each domain, F1 is computed per category and then averaged over the categories (i.e., macro F1).
This yields F1$^H$ for \hab and F1$^R$ for \robo.

\section{Additional Visualization}
\label{appx:moreviz}
\begin{figure}[tp]
    \centering
    \includegraphics[width=\linewidth]{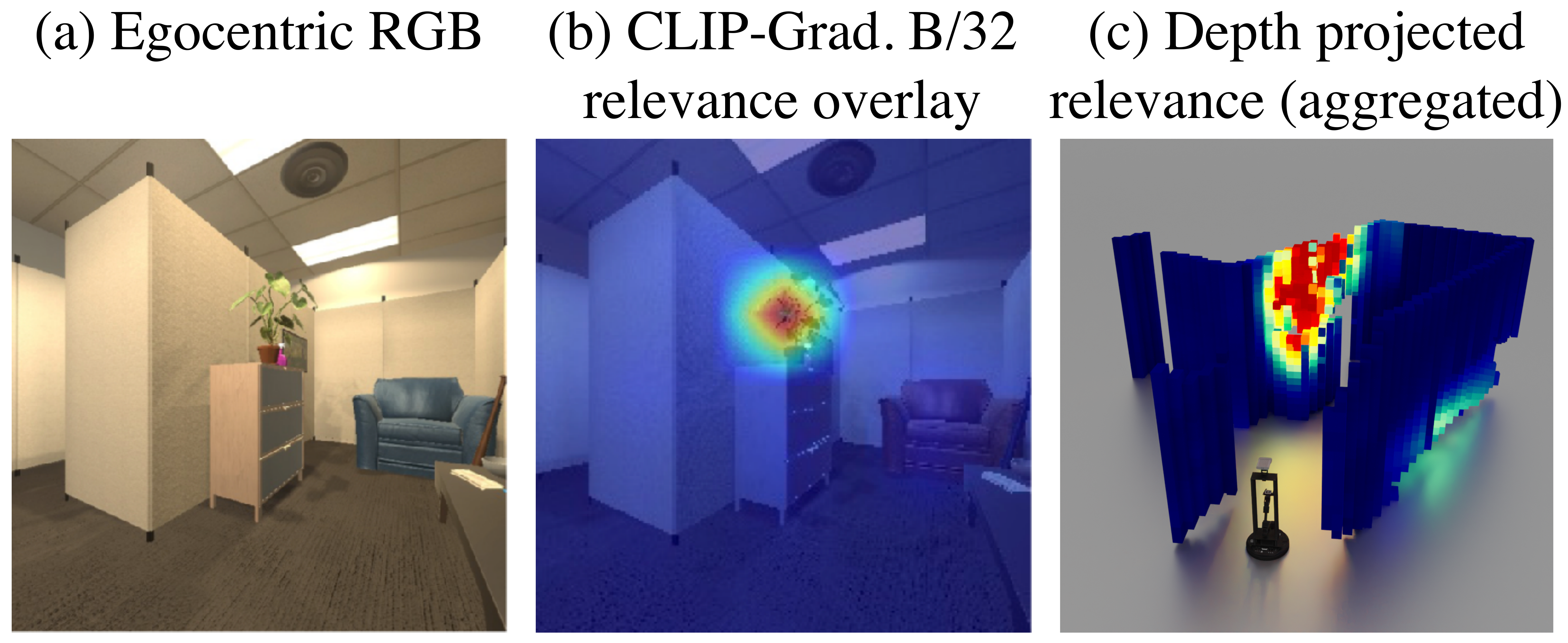}
    \caption{
        \textbf{Projection of object relevance.}
        (a) Egocentric RGB. Note, a CoW also receives a depth image. (b) Raw CLIP-Grad. B/32 prediction for the image targeting the ``plant" class. (c) Back-projection of object relevance, aggregated over time, into a 3D map using agent pose estimates. Areas of high relevence make natural targets for navigation.
    }
    \label{fig:project_viz}
    \vspace{-3mm}
\end{figure}

\begin{figure*}[tp]
    \centering
    \includegraphics[width=\linewidth]{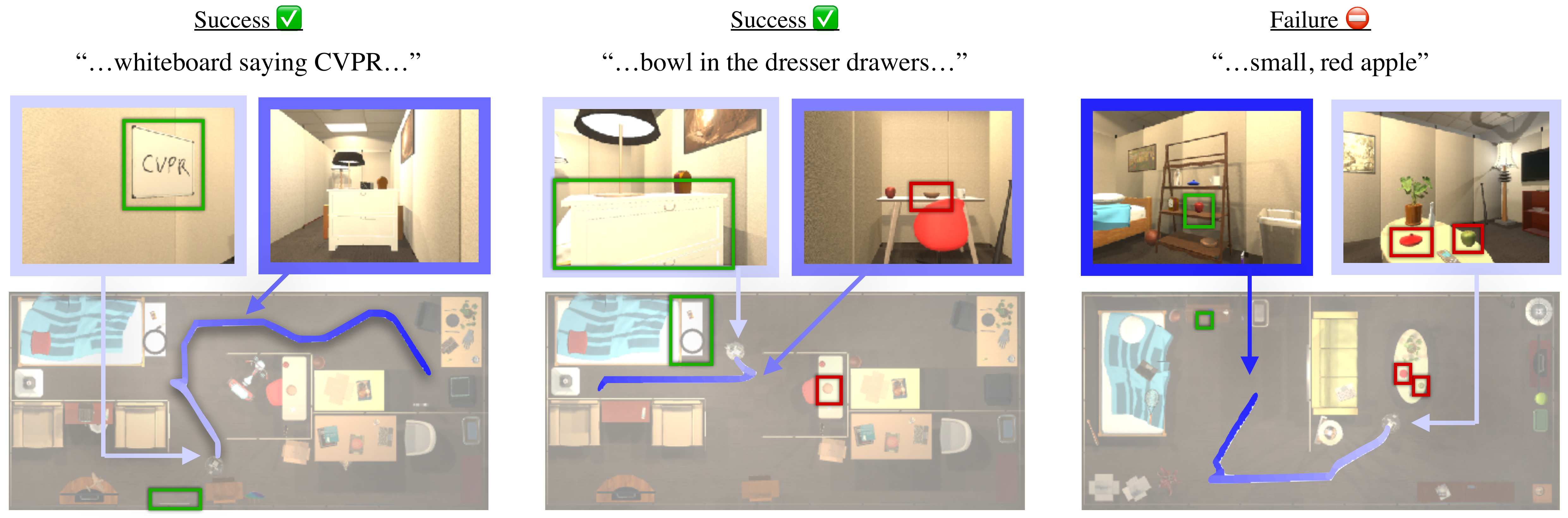}
    \caption{
        \textbf{Trajectory visualization.} Frames are egocentric views. Color indicates trajectory progress, where blue indicating trajectory start and white indicating trajectory end. Target objects are boxed in green, while distractor objects are boxed in red.
    }
    \label{fig:good_viz}
    \vspace{-3mm}
\end{figure*}

To give a more qualitative perspective on our results, we additionally visualize some key aspects of the method and also evaluation trajectories.
In Fig.~\ref{fig:project_viz}, we show back-projecting 2D object relevancy to 3D, which is a key part of the CoW pipeline.
We provide success and failure sample trajectories in Fig.~\ref{fig:good_viz}.

\section{Dataset Details}
\label{appx:dataset}

\begin{figure*}[tp]
    \centering
    \includegraphics[width=0.9\linewidth]{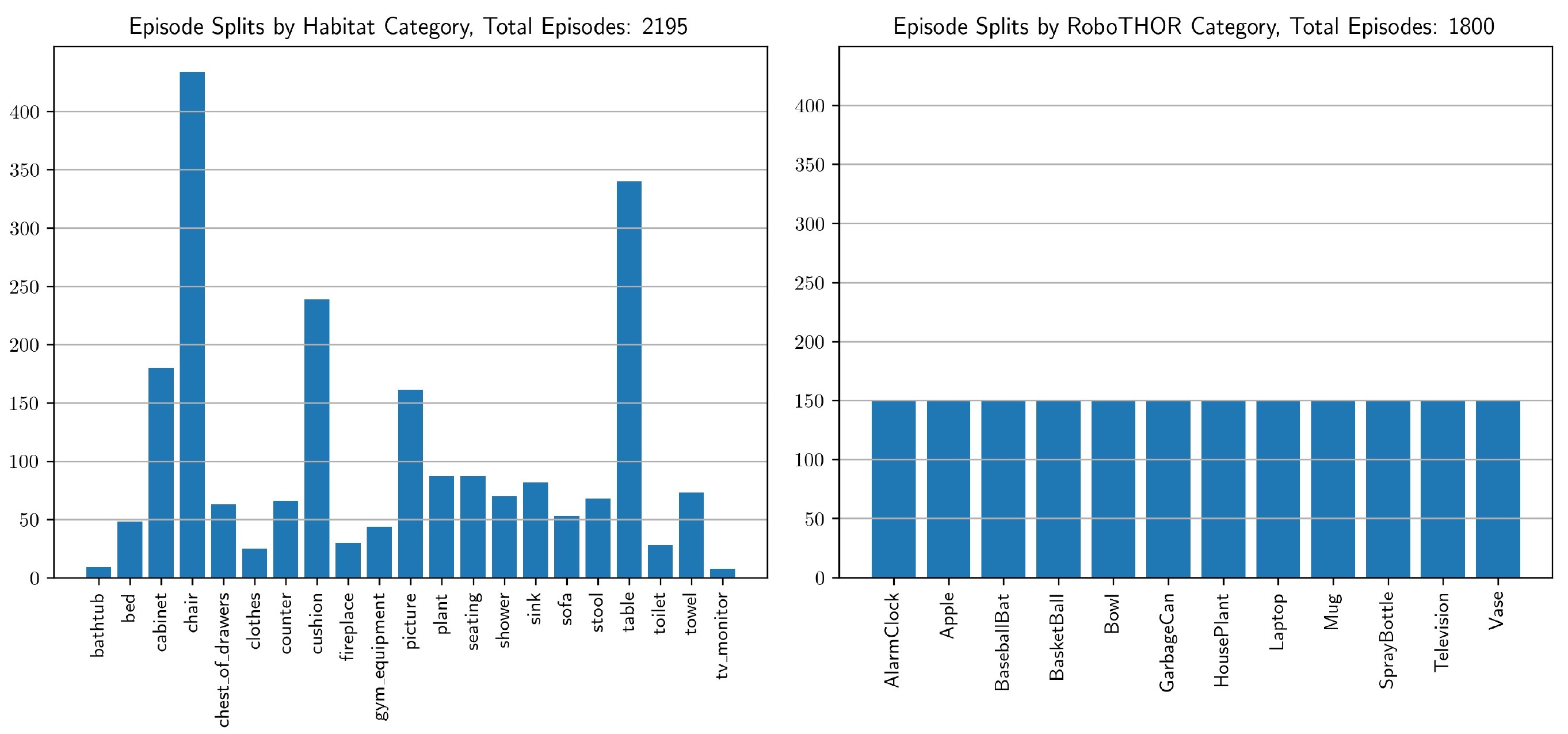}
    \caption{
        \textbf{\hab MP3D and \robo episode splits.} Distribution of episodes in each CVPR 2021 object navigation challenge validation set that we adopt as our test set.
    }
    \label{fig:splits}
    \vspace{-3mm}
\end{figure*}

\mypara{\robo and \hab MP3D dataset details.}
There are 11 \hab MP3D and 15 \robo test scenes, in the official validation sets, which we use as our test sets.
\hab includes 2,195 evaluation episodes, while \robo has 1,800.
We consider the following 21 categories in \hab: 
\texttt{chair},
\texttt{table},
\texttt{picture},
\texttt{cabinet},
\texttt{cushion},
\texttt{sofa},
\texttt{bed},
\texttt{chest\_of\_drawers},
\texttt{plant},
\texttt{sink},
\texttt{toilet},
\texttt{stool},
\texttt{towel},
\texttt{tv\_monitor},
\texttt{shower},
\texttt{bathtub},
\texttt{counter},
\texttt{fireplace},
\texttt{gym\_equipment},
\texttt{seating},
\texttt{clothes}.
We consider the following 12 categories in \robo:
\texttt{AlarmClock},
\texttt{Apple},
\texttt{BaseballBat},
\texttt{BasketBall},
\texttt{Bowl},
\texttt{GarbageCan},
\texttt{HousePlant},
\texttt{Laptop},
\texttt{Mug},
\texttt{SprayBottle},
\texttt{Television},
\texttt{Vase}.
Both of these lists include all objects for the \hab and \robo CVPR 2021 object navigation challenges respectively.
While the distribution of navigation episodes in \robo is equally split per category, this is not the case in \hab MP3D.
We provide testing episode counts per category in Fig.~\ref{fig:splits}.

\begin{figure}[tp]
    \centering
    \includegraphics[width=\linewidth]{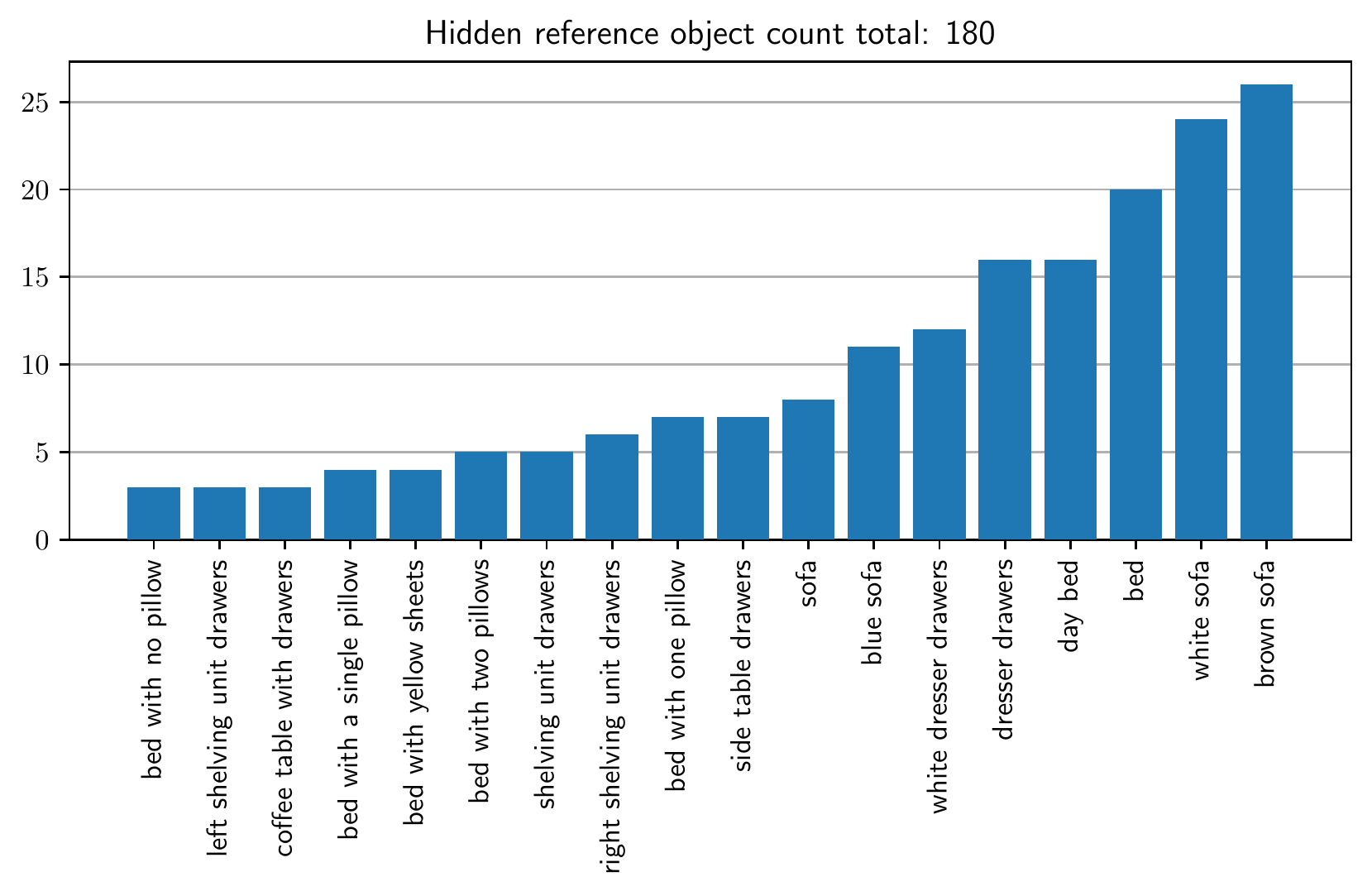}
    \caption{
        \textbf{Hidden reference object distribution for \dataset.}
        Word frequency of large objects that target objects are hidden ``in" or ``under" in the \dataset hidden object captions.
        Names are based on minimal descriptions needed to identify the object in the room.
        For example, ``brown sofa" vs. ``white sofa".
        }
    \label{fig:hidden_freq}
\end{figure}

\mypara{\dataset additional statistics.}
 \begin{figure*}[tp]
    \centering
    \includegraphics[width=\linewidth]{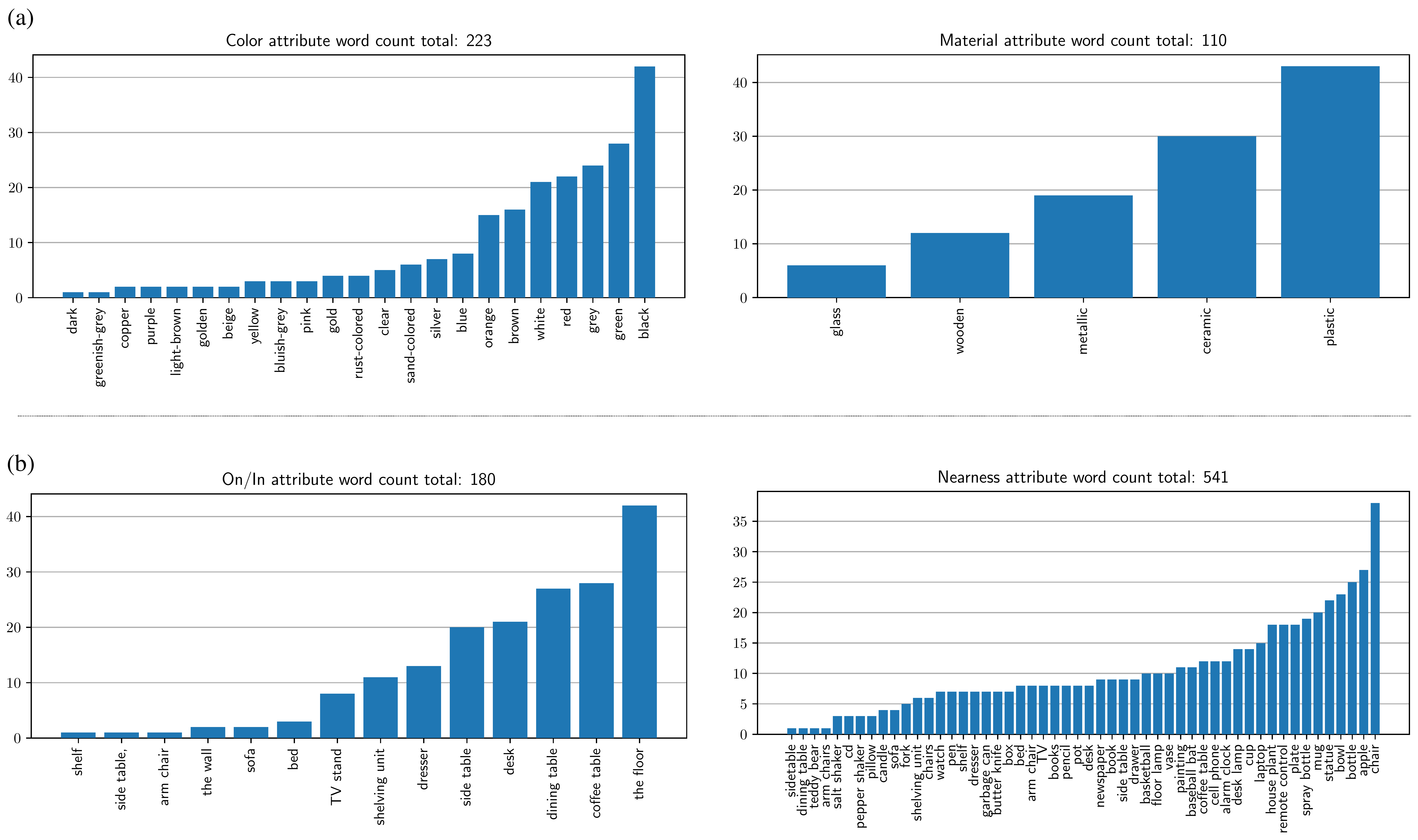}
    \caption{
        \textbf{Attribute distribution for \dataset.}
        (a) Distribution of color and material attribute word frequency in the \dataset appearance captions.
        (b) Distribution of on/in and near attributes by frequency in the \dataset spatial captions.
        }
    \label{fig:added_stats}
\end{figure*} 
In Fig.~\ref{fig:added_stats}, we show word counts for various color, material, and spatial attributes for the appearance and spatial class remapping.
In Fig.~\ref{fig:hidden_freq}, we repeat similar analysis for hidden objects, reporting the word frequency for objects that contain the target objects.

\mypara{\dataset uncommon object asset licensing.}
We found the base instances for the \dataset uncommon object split from CGTrader, an online repository where CAD hobbyists, artists, and professionals host their models.
We choosing our objects, we selected 12 instances that had a ``Royalty Free License" and were free to download at the time of dataset construction.
We acknowledge all artists and provide links to their work:
\begin{itemize}[leftmargin=3mm]
    \item \href{https://www.cgtrader.com/free-3d-models/watercraft/other/surfboard-eadd6608-11e6-4e77-9ae9-74eb52114dad}{``tie-dye surfboard”} by \emph{adhil3dartist} and re-textured to be rainbow
    \vspace{-2mm}
    \item \href{https://www.cgtrader.com/free-3d-models/interior/interior-office/blank-white-board}{``whiteboard saying CVPR"} by \emph{w-stone-art} and re-textured to say CVPR
    \vspace{-2mm}
    \item \href{https://www.cgtrader.com/free-3d-models/household/other/llama-wicker-basket }{``llama wicker basket"} by \emph{eelh}
    \vspace{-2mm}
    \item \href{https://www.cgtrader.com/free-3d-models/household/other/plastic-box-7a6b0bc3-4a54-4744-976a-9097a4fb2677}{``green plastic crate"} by \emph{Snowdrop-2018} and modified to include only the green create
    \vspace{-2mm}
    \item \href{https://www.cgtrader.com/free-3d-models/household/kitchenware/cooker-02}{``rice cooker"} by \emph{fleigh} and re-textured
    \vspace{-2mm}
    \item \href{https://www.cgtrader.com/free-3d-models/food/beverage/argentinian-mate}{``mate gourd"} by \emph{mcgamescompany}
    \vspace{-2mm}
    \item \href{https://www.cgtrader.com/3d-models/vehicle/bicycle/children-bicycle-659a8f0f-99cc-4022-bf15-4aeb308448eb}{``red and blue tricycle"} by \emph{POLY1PROPS}
    \vspace{-2mm}
    \item \href{https://www.cgtrader.com/free-3d-models/electronics/audio/electric-guitar-747681e7-2cc3-4d90-99f7-cfb054c744b3}{``white electric guitar"} by \emph{demolitions2000}
    \vspace{-2mm}
    \item \href{https://www.cgtrader.com/free-3d-models/electronics/other/coffee-maker-2276adc2-c1c7-4054-bf06-1cf673c702ce}{``espresso machine"} by \emph{WolfgangNikolas}
    \vspace{-2mm}
    \item \href{https://www.cgtrader.com/free-3d-models/vehicle/other/free-wooden-airplane-toy}{``wooden toy airplane"} by \emph{fomenos}
    \vspace{-2mm}
    \item \href{https://www.cgtrader.com/free-3d-models/food/miscellaneous/christmas-gingerbread-house}{``gingerbread house"} by \emph{Empire-Assets}
    \vspace{-2mm}
    \item \href{https://www.cgtrader.com/free-3d-models/electronics/computer/rtx-3080ti}{``graphics card"} by \emph{Biggie-3D}
\end{itemize}

\mypara{\dataset sample prompts.}
Here we provide some sample prompts for our appearance, spatial, and hidden object instance remapping.
\begin{itemize}[leftmargin=3mm]
    \item appearance remapping: from ``spray bottle" to ``small, green, plastic spray bottle"
    \vspace{-1mm}
    \item spatial remapping: from ``spray bottle" to ``spray bottle on a coffee table near a house plant"
    \vspace{-1mm}
    \item hidden remapping: from ``spray bottle" to ``spray bottle under the bed"
    \vspace{-1mm}
\end{itemize}

\begin{figure*}[tp]
    \centering
    \includegraphics[width=\linewidth]{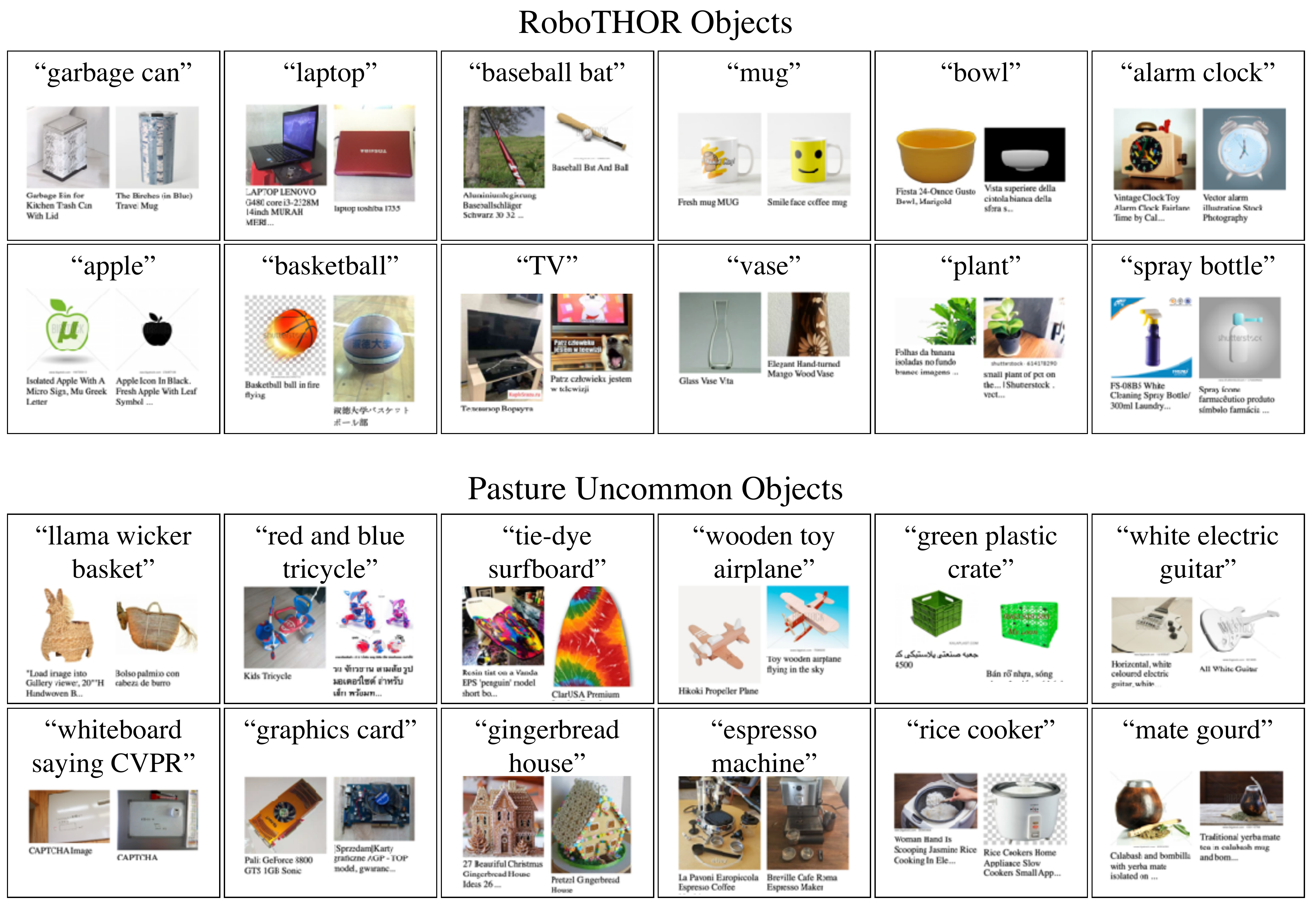}
    \caption{
        \textbf{Qualitative CLIP retrieval.}
        We conduct CLIP retrieval on LAION-5B~\cite{schuhmann2022laion} using the class names shown in quotes with the interface provided \href{https://rom1504.github.io/clip-retrieval/}{here}.
        We show results for both \robo and \dataset uncommon objects.
        While CLIP is not trained on LAION-5B, this figure is included to give an idea of the type of noisy internet image-text data is trained on.
        Retrieval returns reasonable results for uncommon objects, suggesting CLIP is able to semantically distinguish these objects.
        }
    \label{fig:retrieval}
\end{figure*}

\mypara{Retrieval of uncommon objects on LAION-5B~\cite{schuhmann2022laion}}
We include sample data from CLIP retrieval for both \robo and \dataset uncommon objects in Fig.~\ref{fig:retrieval}.
We included the results to show that CLIP is familiar with the concepts that we chose when creating the uncommon split of \dataset and that qualitative image results rival those of the more common \robo objects.

\section{Prompt Ensemble Ablation}
\label{appx:prompt_ablation}

\begin{table}[tp]
\centering
\tabcolsep=0.23cm
\scriptsize
\begin{tabular}{cccc|l|yc}
\toprule
\multicolumn{5}{c|}{CoW breeds} & \multicolumn{2}{c}{\robo}\\
ID & Loc. & Arch. & Post  & Exp. Strategy  & \textsc{SPL} & \textsc{SR}\\\midrule
\textcolor{tab:purple}{\Large{$\triangle$}} & CLIP-Grad. & B/32 & \checkmark  & photo & 9.5 & \textbf{15.2}\\
\textcolor{tab:purple}{\Large{$\triangle$}} & CLIP-Grad. & B/32 & \checkmark  & prompt ens. & \textbf{9.7} & \textbf{15.2}\\
\bottomrule

\end{tabular}
\vspace*{-1mm}
\caption{\textbf{Prompt ablation.} For a fixed object localizers (CLIP-Grad. B/32 with post processing), we ablate over different choices of prompts. We find that the 80 prompt ensemble (prompt ens.), introduced by OpenAI, outperforms the simple prompt: ``a photo of a \{\}." (photo) in most cases. However, deltas are not large, suggesting that this is a less critical design decision in the CoW framework.
}
\label{tab:prompt}
\vspace{-5mm}
\end{table}
To verify the benefits of using the 80 prompt ensemble created by OpenAI---denoted as \emph{prompt ens.}---, we compare performance for two models on \robo.
The competing approach uses a single prompt for each class ``a photo of a \{\}."---denoted as \emph{photo}.
As we see in Tab.~\ref{tab:prompt}, prompt ens. boosts \textsc{SPL} by by 0.2 for the CLIP-Grad. model.
These, results suggest that the two variations of prompting strategies have marginal influence on downstream performance.
For all experiments in the main paper that use CLIP, we use the 80 prompt ensemble in conjunction with the class label specific for a task (e.g., ``spray bottle under the bed").

\section{Category-level Results}
\label{appx:cat}

\begin{table}
\centering
\small\tabcolsep=0.09cm
\begin{tabular}{l|cycy|cycy}
\toprule
 & \multicolumn{4}{c|}{\dataset} & \multicolumn{4}{c}{\dataset}\\
 & \multicolumn{4}{c|}{Appear.} & \multicolumn{4}{c}{Space}\\
 & \multicolumn{2}{c}{\textcolor{tab:purple}{\Large{$\triangle$}}} & \multicolumn{2}{c|}{\textcolor{tab:cyan}{\Large{$\triangle$}}} & \multicolumn{2}{c}{\textcolor{tab:purple}{\Large{$\triangle$}}} & \multicolumn{2}{c}{\textcolor{tab:cyan}{\Large{$\triangle$}}}\\
category & \textsc{{SR}} & \textsc{{SPL}} & \textsc{{SR}} & \textsc{{SPL}} & \textsc{{SR}} & \textsc{{SPL}} & \textsc{{SR}} & \textsc{{SPL}}\\\midrule
\textsc{AlarmClock}& 6.7& 4.0& \textbf{23.3}& \textbf{10.7}& 3.3& 3.3& \textbf{10.0}& \textbf{3.8}\\
\textsc{Apple}& 6.7& 5.7& \textbf{36.7}& \textbf{17.0}& \textbf{10.0}& \textbf{8.4}& 3.3& 1.0\\
\textsc{BaseballBat}& 0.0& 0.0& \textbf{3.3}& \textbf{1.2}& 3.3& 2.5& \textbf{6.7}& \textbf{2.7}\\
\textsc{BasketBall}& 6.7& 2.8& \textbf{36.7}& \textbf{24.1}& 10.0& 5.6& \textbf{36.7}& \textbf{26.8}\\
\textsc{Bowl}& 3.3& 0.5& \textbf{13.3}& \textbf{5.9}& 10.0& 5.6& \textbf{16.7}& \textbf{6.9}\\
\textsc{GarbageCan}& 26.7& 20.2& \textbf{50.0}& \textbf{31.5}& 30.0& 23.0& \textbf{40.0}& \textbf{23.2}\\
\textsc{HousePlant}& 20.0& 16.9& \textbf{30.0}& \textbf{20.2}& 13.3& 10.8& \textbf{40.0}& \textbf{21.9}\\
\textsc{Laptop}& 13.3& 10.6& \textbf{20.0}& \textbf{11.5}& 13.3& 9.6& \textbf{20.0}& \textbf{13.7}\\
\textsc{Mug}& 10.0& 7.5& \textbf{46.7}& \textbf{27.4}& 10.0& \textbf{7.5}& \textbf{13.3}& 5.4\\
\textsc{SprayBottle}& 16.7& 13.6& \textbf{33.3}& \textbf{19.2}& \textbf{16.7}& \textbf{15.8}& \textbf{16.7}& 6.8\\
\textsc{Television}& 10.0& \textbf{10.0}& \textbf{13.3}& 8.9& 6.7& 6.4& \textbf{20.0}& \textbf{9.9}\\
\textsc{Vase}& \textbf{23.3}& \textbf{17.5}& 10.0& 9.1& \textbf{13.3}& \textbf{9.8}& 10.0& 5.4\\

\bottomrule
\end{tabular}
\caption{\textbf{Attribute object navigation.}
Appearance-based captions consistently perform better than spatial captions. OWL consistently performs better than CLIP-Grad.
}
\label{tab:category_attr}
\end{table}

\begin{table}
\centering
\small\tabcolsep=0.09cm
\begin{tabular}{l|cycy|cycy}
\toprule
 & \multicolumn{4}{c|}{\dataset} & \multicolumn{4}{c}{\dataset}\\
 & \multicolumn{4}{c|}{Appear. distract} & \multicolumn{4}{c}{Space distract}\\
 & \multicolumn{2}{c}{\textcolor{tab:purple}{\Large{$\triangle$}}} & \multicolumn{2}{c|}{\textcolor{tab:cyan}{\Large{$\triangle$}}} & \multicolumn{2}{c}{\textcolor{tab:purple}{\Large{$\triangle$}}} & \multicolumn{2}{c}{\textcolor{tab:cyan}{\Large{$\triangle$}}}\\
category & \textsc{{SR}} & \textsc{{SPL}} & \textsc{{SR}} & \textsc{{SPL}} & \textsc{{SR}} & \textsc{{SPL}} & \textsc{{SR}} & \textsc{{SPL}}\\\midrule
\textsc{AlarmClock}& 3.3& 3.0& \textbf{13.3}& \textbf{6.5}& \textbf{6.7}& \textbf{6.3}& \textbf{6.7}& 2.8\\
\textsc{Apple}& \textbf{10.0}& 6.4& \textbf{10.0}& \textbf{7.4}& 3.3& 3.3& \textbf{10.0}& \textbf{4.0}\\
\textsc{BaseballBat}& 0.0& 0.0& \textbf{13.3}& \textbf{4.5}& 0.0& 0.0& \textbf{10.0}& \textbf{8.1}\\
\textsc{BasketBall}& 6.7& 3.3& \textbf{20.0}& \textbf{12.6}& \textbf{16.7}& 9.4& \textbf{16.7}& \textbf{9.7}\\
\textsc{Bowl}& 3.3& 3.2& \textbf{16.7}& \textbf{8.5}& 10.0& 8.6& \textbf{23.3}& \textbf{12.6}\\
\textsc{GarbageCan}& 26.7& 19.9& \textbf{30.0}& \textbf{21.6}& 13.3& 10.5& \textbf{26.7}& \textbf{18.2}\\
\textsc{HousePlant}& 10.0& 6.0& \textbf{16.7}& \textbf{11.0}& 13.3& 10.8& \textbf{23.3}& \textbf{13.7}\\
\textsc{Laptop}& 16.7& \textbf{13.6}& \textbf{23.3}& 11.9& \textbf{16.7}& \textbf{11.9}& \textbf{16.7}& 11.8\\
\textsc{Mug}& 6.7& 5.1& \textbf{26.7}& \textbf{17.8}& \textbf{10.0}& \textbf{7.8}& 6.7& 2.7\\
\textsc{SprayBottle}& 13.3& 12.4& \textbf{26.7}& \textbf{15.2}& 16.7& \textbf{15.4}& \textbf{20.0}& 8.0\\
\textsc{Television}& 6.7& 6.6& \textbf{26.7}& \textbf{15.5}& 6.7& 3.3& \textbf{20.0}& \textbf{12.8}\\
\textsc{Vase}& \textbf{13.3}& \textbf{10.2}& 10.0& 8.6& 10.0& 6.5& \textbf{13.3}& \textbf{8.4}\\

\bottomrule
\end{tabular}
\caption{\textbf{Attribute object navigation with distractors.}
Distractors consistently hurt performance compared to the no distractor numbers in Tab.~\ref{tab:category_attr}, suggesting that models cannot make full use of remapped classes with attributes.
}
\label{tab:category_attr_dist}
\end{table}

\begin{table}
\centering
\small\tabcolsep=0.09cm
\begin{tabular}{l|cycy|cycy}
\toprule
 & \multicolumn{4}{c|}{\dataset} & \multicolumn{4}{c}{\dataset}\\
 & \multicolumn{4}{c|}{Hidden} & \multicolumn{4}{c}{Hidden distract}\\
 & \multicolumn{2}{c}{\textcolor{tab:purple}{\Large{$\triangle$}}} & \multicolumn{2}{c|}{\textcolor{tab:cyan}{\Large{$\triangle$}}} & \multicolumn{2}{c}{\textcolor{tab:purple}{\Large{$\triangle$}}} & \multicolumn{2}{c}{\textcolor{tab:cyan}{\Large{$\triangle$}}}\\
category & \textsc{{SR}} & \textsc{{SPL}} & \textsc{{SR}} & \textsc{{SPL}} & \textsc{{SR}} & \textsc{{SPL}} & \textsc{{SR}} & \textsc{{SPL}}\\\midrule
\textsc{AlarmClock}& \textbf{26.7}& \textbf{15.8}& 6.7& 3.0& 20.0& \textbf{9.9}& \textbf{23.3}& 8.3\\
\textsc{Apple}& 10.0& \textbf{9.5}& \textbf{13.3}& 8.9& 6.7& 4.7& \textbf{10.0}& \textbf{7.9}\\
\textsc{BaseballBat}& \textbf{13.3}& \textbf{9.2}& \textbf{13.3}& 5.8& \textbf{10.0}& \textbf{7.2}& 3.3& 2.9\\
\textsc{BasketBall}& 10.0& 5.1& \textbf{26.7}& \textbf{15.2}& 16.7& 12.2& \textbf{20.0}& \textbf{15.0}\\
\textsc{Bowl}& 16.7& 9.2& \textbf{23.3}& \textbf{11.9}& \textbf{23.3}& \textbf{15.1}& 20.0& 10.6\\
\textsc{GarbageCan}& 13.3& 7.8& \textbf{26.7}& \textbf{13.8}& 13.3& 7.0& \textbf{16.7}& \textbf{10.9}\\
\textsc{HousePlant}& 13.3& 9.6& \textbf{16.7}& \textbf{10.8}& \textbf{16.7}& 10.9& \textbf{16.7}& \textbf{11.6}\\
\textsc{Laptop}& \textbf{10.0}& 8.1& \textbf{10.0}& \textbf{8.4}& \textbf{26.7}& \textbf{20.7}& 10.0& 8.4\\
\textsc{Mug}& 13.3& 7.3& \textbf{33.3}& \textbf{23.8}& 20.0& 12.5& \textbf{23.3}& \textbf{18.4}\\
\textsc{SprayBottle}& \textbf{13.3}& \textbf{8.4}& 0.0& 0.0& \textbf{20.0}& \textbf{10.3}& 0.0& 0.0\\
\textsc{Television}& 16.7& 8.0& \textbf{36.7}& \textbf{22.9}& 10.0& 6.9& \textbf{16.7}& \textbf{11.4}\\
\textsc{Vase}& 16.7& \textbf{11.9}& \textbf{23.3}& 11.2& 10.0& 7.0& \textbf{13.3}& \textbf{7.0}\\

\bottomrule
\end{tabular}
\caption{\textbf{Hidden object navigation category-level results.}
}
\label{tab:category_hidden}
\end{table}

\begin{table}
\centering
\small
\begin{tabular}{l|cycy}
\toprule
 & \multicolumn{4}{c}{\dataset}\\
 & \multicolumn{4}{c}{Uncom.}\\
 & \multicolumn{2}{c}{\textcolor{tab:purple}{\Large{$\triangle$}}} & \multicolumn{2}{c}{\textcolor{tab:cyan}{\Large{$\triangle$}}}\\
category & \textsc{{SR}} & \textsc{{SPL}} & \textsc{{SR}} & \textsc{{SPL}}\\\midrule
\textsc{GingerbreadHouse}& 20.0& 14.3& \textbf{26.7}& \textbf{18.6}\\
\textsc{EspressoMachine}& 10.0& 7.7& \textbf{46.7}& \textbf{24.6}\\
\textsc{Crate}& 23.3& 18.2& \textbf{40.0}& \textbf{27.0}\\
\textsc{ElectricGuitar}& 16.7& 10.0& \textbf{46.7}& \textbf{30.8}\\
\textsc{RiceCooker}& 3.3& 2.9& \textbf{20.0}& \textbf{11.6}\\
\textsc{LlamaWickerBasket}& 16.7& 12.6& \textbf{30.0}& \textbf{24.5}\\
\textsc{Whiteboard}& \textbf{63.3}& \textbf{43.2}& 30.0& 18.7\\
\textsc{Surfboard}& 26.7& 20.6& \textbf{60.0}& \textbf{38.9}\\
\textsc{Tricycle}& 10.0& 9.0& \textbf{53.3}& \textbf{31.7}\\
\textsc{GraphicsCard}& 3.3& 2.1& \textbf{13.3}& \textbf{6.0}\\
\textsc{Mate}& \textbf{0.0}& \textbf{0.0}& \textbf{0.0}& \textbf{0.0}\\
\textsc{ToyAirplane}& 0.0& 0.0& \textbf{26.7}& \textbf{13.7}\\

\bottomrule
\end{tabular}
\caption{\textbf{Uncommon object navigation category-level results.}}
\label{tab:category_uncom}
\end{table}

\begin{table}
\centering
\small
\begin{tabular}{l|cycy}
\toprule
 & \multicolumn{4}{c}{\robo}\\
 & \multicolumn{2}{c}{\textcolor{tab:purple}{\Large{$\triangle$}}} & \multicolumn{2}{c}{\textcolor{tab:cyan}{\Large{$\triangle$}}}\\
category & \textsc{{SR}} & \textsc{{SPL}} & \textsc{{SR}} & \textsc{{SPL}}\\\midrule
\textsc{AlarmClock}& 5.3& 3.1& \textbf{30.7}& \textbf{19.2}\\
\textsc{Apple}& 15.3& 9.7& \textbf{34.0}& \textbf{19.6}\\
\textsc{BaseballBat}& \textbf{4.0}& \textbf{1.5}& 2.0& 0.5\\
\textsc{BasketBall}& 19.3& 14.8& \textbf{36.0}& \textbf{25.3}\\
\textsc{Bowl}& 5.3& 4.0& \textbf{18.0}& \textbf{11.1}\\
\textsc{GarbageCan}& 30.0& 21.0& \textbf{50.0}& \textbf{32.2}\\
\textsc{HousePlant}& 30.7& 18.7& \textbf{36.7}& \textbf{24.8}\\
\textsc{Laptop}& 16.0& 10.6& \textbf{20.0}& \textbf{11.3}\\
\textsc{Mug}& 10.7& 5.0& \textbf{40.7}& \textbf{25.5}\\
\textsc{SprayBottle}& 8.0& 5.2& \textbf{23.3}& \textbf{13.8}\\
\textsc{Television}& \textbf{29.3}& \textbf{16.9}& 23.3& 13.4\\
\textsc{Vase}& \textbf{8.0}& \textbf{5.8}& 6.0& 5.7\\

\bottomrule
\end{tabular}
\caption{\textbf{\robo category-level results.}}
\label{tab:category_robo}
\end{table}

\begin{table}
\centering
\small
\begin{tabular}{l|cycy}
\toprule
 & \multicolumn{4}{c}{\dataset}\\
 & \multicolumn{4}{c}{\hab}\\
 & \multicolumn{2}{c}{\textcolor{tab:purple}{\Large{$\triangle$}}} & \multicolumn{2}{c}{\textcolor{tab:cyan}{\Large{$\triangle$}}}\\
category & \textsc{{SR}} & \textsc{{SPL}} & \textsc{{SR}} & \textsc{{SPL}}\\\midrule
\textsc{chair}& 5.1& 2.2& \textbf{7.4}& \textbf{3.7}\\
\textsc{table}& \textbf{34.7}& \textbf{18.4}& 21.2& 9.6\\
\textsc{picture}& 1.2& 0.5& \textbf{1.9}& \textbf{1.1}\\
\textsc{cabinet}& \textbf{9.4}& 4.7& 8.3& \textbf{4.9}\\
\textsc{cushion}& 5.0& 3.0& \textbf{12.1}& \textbf{5.8}\\
\textsc{sofa}& \textbf{0.0}& \textbf{0.0}& \textbf{0.0}& \textbf{0.0}\\
\textsc{bed}& \textbf{0.0}& \textbf{0.0}& \textbf{0.0}& \textbf{0.0}\\
\textsc{chest\_of\_drawers}& 0.0& 0.0& \textbf{1.6}& \textbf{0.4}\\
\textsc{plant}& \textbf{5.7}& \textbf{3.6}& 2.3& 1.3\\
\textsc{sink}& \textbf{2.4}& 1.6& \textbf{2.4}& \textbf{1.8}\\
\textsc{toilet}& \textbf{0.0}& \textbf{0.0}& \textbf{0.0}& \textbf{0.0}\\
\textsc{stool}& 0.0& 0.0& \textbf{2.9}& \textbf{2.4}\\
\textsc{towel}& 0.0& 0.0& \textbf{1.4}& \textbf{0.7}\\
\textsc{tv\_monitor}& \textbf{0.0}& \textbf{0.0}& \textbf{0.0}& \textbf{0.0}\\
\textsc{shower}& \textbf{7.1}& \textbf{4.4}& 1.4& 0.9\\
\textsc{bathtub}& 0.0& 0.0& \textbf{11.1}& \textbf{3.1}\\
\textsc{counter}& \textbf{6.1}& \textbf{3.1}& 0.0& 0.0\\
\textsc{fireplace}& \textbf{10.0}& \textbf{5.0}& 3.3& 1.7\\
\textsc{gym\_equipment}& \textbf{0.0}& \textbf{0.0}& \textbf{0.0}& \textbf{0.0}\\
\textsc{seating}& \textbf{12.6}& \textbf{5.2}& 0.0& 0.0\\
\textsc{clothes}& \textbf{8.0}& \textbf{4.3}& 4.0& 2.1\\

\bottomrule
\end{tabular}
\caption{\textbf{\hab category-level results.}}
\label{tab:category_hab}
\end{table}

For completeness we also include salient category-level results for an OWL (\textcolor{tab:cyan}{$\triangle$}) and CLIP-Grad. (\textcolor{tab:purple}{$\triangle$}) B/32 models with post-processing for \hab MP3D, \robo, and \dataset.
For a comparison for the appearance (Appear.) and spatial (Space) \dataset splits, see Tab.~\ref{tab:category_attr}.
For Appear. and Space with distractors (distract), see Tab.~\ref{tab:category_attr_dist}.
For hidden object with and without distractors see Tab.~\ref{tab:category_hidden}.
For uncommon objects see Tab.~\ref{tab:category_uncom}.
For \robo see Tab.~\ref{tab:category_robo}.
For \hab see Tab.~\ref{tab:category_hab}.

\section{Additional Failure Analysis}
\label{appx:fail}
\begin{figure*}[tp]
    \centering
    \includegraphics[width=\linewidth]{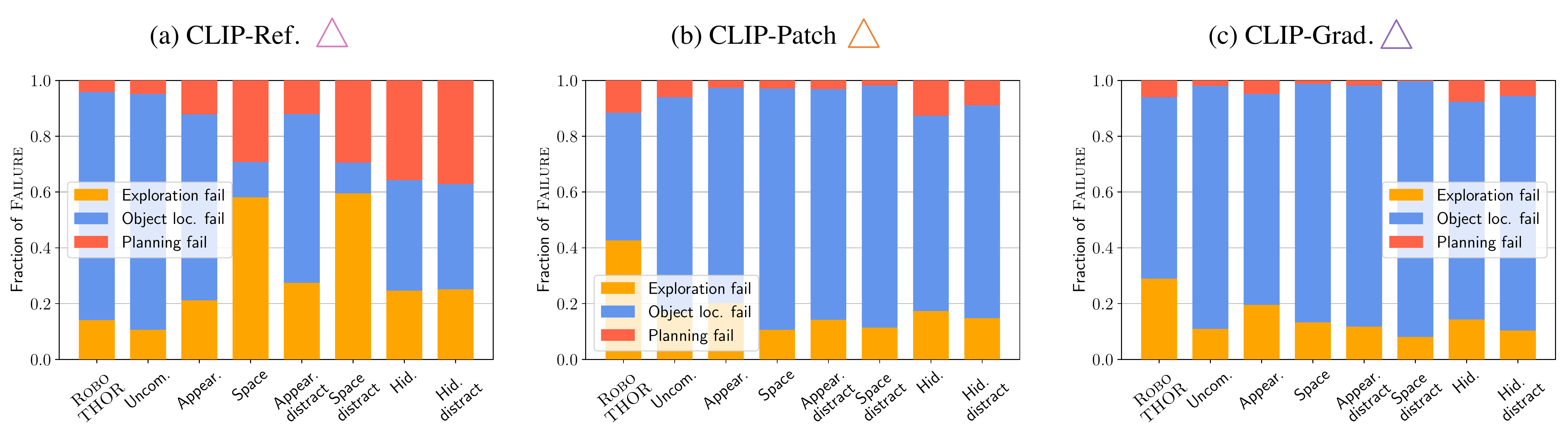}
    \caption{
        \textbf{Failure analysis for more models.}
        (a, b, c) All show additional error analysis. When comparing CLIP-Patch and CLIP-Grad. to OWL shown in Fig.~\ref{fig:fail}, we notice that the former have a higher percentage of object localization failures and also lower success rate downstream on average.
    }
    \label{fig:faul_plus_plus}
    \vspace{-3mm}
\end{figure*}

For addition failure analysis on CLIP-Ref., CLIP-Patch, and CLIP-Grad. see Fig.~\ref{fig:faul_plus_plus}.
All models have a ViT-B/32 architecture and apply post-processing.
We notice that that CLIP-Patch and CLIP-Grad. have a higher fraction of object localization failure when compared to OWL.
For CLIP-Ref., where performance is in the very low success regime (e.g., $<3\%$), we notice less consistent patterns.

\end{document}